\documentclass[mathematics,article,submit,moreauthors,pdftex]{Definitions/mdpi}

\newcolumntype{Y}{>{\centering\arraybackslash}X}
\newcolumntype{C}[1]{>{\centering\arraybackslash}m{#1}} 

\definecolor{ceruleanblue}{rgb}{0.16, 0.32, 0.75}

\firstpage{1} 
\pubvolume{9}
\issuenum{23}
\articlenumber{3137}
\doinum{10.3390/math9233137}
\pubyear{2021}
\copyrightyear{2021}
\externaleditor{{'}
 } 
\datereceived{} 
\dateaccepted{} 
\datepublished{} 
\hreflink{https://doi.org/10.3390/math9233137}
\pdfoutput=1

\Title{XCM: An Explainable Convolutional Neural Network for Multivariate Time Series Classification}

\TitleCitation{XCM: An Explainable Convolutional Neural Network for Multivariate Time Series Classification}

\Author{Kevin Fauvel 
 $^{1,}$*, Tao Lin $^{2}$, V\'e{ronique Masson} 
 $^{3}$, \'E{lisa Fromont}
 $^{4}$ and Alexandre Termier $^{3}$}
\AuthorCitation{Fauvel, K., Lin, T.; Masson, V.; Fromont, \'E.; Termier, A.}
\address{$^{1}$ \quad Inria, Univ Rennes, CNRS, IRISA, Rennes, France;\\
$^{2}$ \quad Zhejiang University, Hangzhou, China;\\
$^{3}$ \quad Univ Rennes, Inria, CNRS, IRISA, Rennes, France;\\
$^{4}$ \quad Univ Rennes, IUF, Inria, CNRS, IRISA, Rennes, France; \\
kevin.fauvel@inria.fr (K.F.); lintao1@zju.edu.cn (T.L.); veronique.masson@irisa.fr (V.M.); elisa.fromont@irisa.fr\\ (E.F.); alexandre.termier@irisa.fr (A.T.)\\
}
\corres{Correspondence: kevin.fauvel@inria.fr}

\abstract{
Multivariate Time Series (MTS) classification has gained importance over the past decade with the increase in the number of temporal datasets in multiple domains. The current state-of-the-art MTS classifier is a heavyweight deep learning approach, which outperforms the second-best MTS classifier only on large datasets. Moreover, this deep learning approach cannot provide faithful explanations as it relies on post hoc model-agnostic explainability methods, which could prevent its use in numerous applications.
In this paper, we present XCM, an eXplainable Convolutional neural network for MTS classification. 
XCM is a new compact convolutional neural network which extracts information relative to the observed variables and time directly from the input data.
Thus, XCM architecture enables a good generalization ability on both large and small datasets, while allowing the full exploitation of a faithful post hoc model-specific explainability method (Gradient-weighted Class Activation Mapping) by precisely identifying the observed variables and timestamps of the input data that are important for predictions.
We first show that XCM outperforms the state-of-the-art MTS classifiers on both the large and small public UEA datasets.
Then, we illustrate how XCM reconciles performance and explainability on a synthetic dataset and show that XCM enables a more precise identification of the regions of the input data that are important for predictions compared to the current deep learning MTS classifier also providing faithful explainability.
Finally, we present how XCM can outperform the current most accurate state-of-the-art algorithm on a real-world application while enhancing explainability by providing faithful and more informative explanations.
}

\keyword{convolutional neural network; explainability; multivariate time series classification} 

\begin{document}

\section{Introduction}
Following the remarkable availability of multivariate temporal data, Multivariate Time Series (MTS) analysis is becoming a necessary procedure in a wide range of application domains (e.g., {finance~\citep{Chen19},} healthcare~\citep{Li18}, mobility~\citep{Jiang19}, and natural disasters~\citep{Fauvel20}). A~time series is a sequence of real values ordered according to time; and when a set of coevolving time series are recorded simultaneously by a set of sensors, it is called an MTS. In~this paper, we address the issue of MTS classification, which consists of learning the relationship between an MTS and its~label. 

According to the results published, the~most accurate state-of-the-art MTS classifier on average is a deep learning approach (MLSTM-FCN~\citep{Karim19}). MLSTM-FCN consists of the concatenation of a Long Short-Term Memory (LSTM) block with a Convolutional Neural Network (CNN) block composed of three convolutional sub-blocks. However, MLSTM-FCN outperforms the second-best MTS classifier (Bag-of-Words method WEASEL+MUSE~\citep{Schafer17}) only on the large datasets (relatively to the public UEA archive~\citep{Bagnall18}---training set \mbox{size $\geq$ 500}). This deep learning approach contains a significant number of trainable parameters, which could be an important reason for its poor performance on small datasets. Moreover, for~many applications, the~adoption of machine learning methods cannot rely solely on their prediction performance. For~example, the~European Union's General Data Protection Regulation (GDPR), which became enforceable on 25 May 2018, introduces a right to explanation for all individuals so that they can obtain ``meaningful explanations of the logic involved'' when automated decision making has ``legal effects'' on individuals or similarly ``significantly affecting'' them  {(\url{https://ec.europa.eu/info/law/law-topic/data-protection_en} (accessed on 1 November 2021))}.
{As far as we have seen, an~architecture concatenating a LSTM network with a CNN such as MLSTM-FCN, or~a classifier based on unigrams/bigrams extraction following a Symbolic Fourier Approximation~\citep{Schafer12} such as WEASEL+MUSE, cannot provide perfectly faithful explanations as they rely solely on post hoc model-agnostic explainability methods~\citep{Rudin19}, which could prevent their use in numerous applications.}
Faithfulness is critical as it corresponds to the level of trust an end-user can have in the explanations of model predictions, i.e.,~the level of relatedness of the explanations to what the model actually computes.
Hence, we propose a new compact (in terms of the number of parameters) and explainable deep learning approach for MTS classification that performs well on both large and small datasets while providing faithful~explanations.

CNNs along with post hoc model-specific saliency methods such as Gradient-weighted Class Activation Mapping---Grad-CAM~\citep{Selvaraju19}---have the potential to have a compact architecture while enabling faithful explanations~\citep{Adebayo18}. A~recent CNN, MTEX-CNN~\citep{Assaf19} proposes using 2D and 1D convolution filters in sequence to extract key MTS information, i.e.,~information relative to the observed variables and time, respectively.
However, as~confirmed by our experiments, the~features related to time which are extracted from the output features of the first stage (relative to each observed variable) cannot fully incorporate the timing information from the input data and~subsequently yield poor performance compared to the state-of-the-art MTS classifiers.
In addition, the~significant number of trainable parameters of MTEX-CNN affects its generalization ability on small datasets.
Finally, MTEX-CNN requires upsampling processes on feature maps when applying Grad-CAM, which can lead to an imprecise identification of the regions of the input data that are important for~predictions.

Therefore, we propose a new faithfully eXplainable CNN method for MTS classification (XCM) which improves MTEX-CNN in three substantial ways: \textit{(i)} it generates features by extracting information relative to the observed variables and timestamps in parallel and directly from the input data; \textit{(ii)} it enhances the generalization ability by adopting a compact architecture (in terms of the number of parameters); and \textit{(iii)} it allows precise identification of the observed variables and timestamps of the input data that are important for predictions by avoiding upsampling processes.
Summarizing our main contributions:

\begin{itemize}
	\item We present XCM, an~end-to-end new compact and explainable convolutional neural network for MTS classification which supports its predictions with faithful~explanations;
	\item We show that XCM outperforms the state-of-the-art MTS classifiers on both the large and small UEA datasets~\citep{Bagnall18};
	\item We illustrate on a synthetic dataset that XCM enables a more precise identification of the regions of the input data that are important for predictions compared to the current faithfully explainable deep learning MTS classifier MTEX-CNN;
	\item We show that XCM outperforms the current most accurate state-of-the-art algorithm on a real-world application while enhancing explainability by providing faithful and more informative explanations.
\end{itemize}

The rest of this paper is organized as follows: Section~\ref{sec2} presents the related work concerning MTS classification and explainability; Section~\ref{sec3} details XCM architecture; Section~\ref{sec4} presents our evaluation method; and finally, Section~\ref{XCM_results} discusses our results.

\section{Related~Work} \label{sec2}
In this section, we first introduce the background of our study. Then, we present the state-of-the-art MTS classifiers, and~we end with existing explainability methods supporting CNNs models'~predictions.

\subsection{Background}
{We address the issue of supervised learning for classification. \textit{Classification} consists of learning a function that maps an input data to its label: given an input space $X$, an~output space $Y$, an~unknown distribution ${P}$ over $X \times Y$, a~training set sampled from ${P}$, and a~0--1 loss function $\ell_{0-1}$ compute function $h^*$ as follows:
}
\begin{equation}
	{h^* = \underset{h}{\arg\min} \quad \mathbb{E}_{(x,y) \sim P} \left[ \ell_{0-1}(h, (x,y)) \right]}
\end{equation}

In this study, classification is performed on multivariate time series datasets. A~\textit{Multivariate Time Series} (MTS) $M=\{x_1,...,x_d\} \in \mathcal{R}^{d*l}$ is an ordered sequence of $d \in \mathcal{N}$ streams with $x_i=(x_{i,1},...,x_{i,l})$, where $l$ is the length of the time series and $d$ is the number of multivariate dimensions. We address MTS generated from automatic sensors with a fixed and synchronized sampling along all dimensions. An~example of an MTS with two dimensions and a length of 100 is given at the top of Figure~\ref{fig:synthetic_MTEX_XCM}.

Before presenting the state-of-the-art MTS classifiers, we introduce some notions about neural networks and the subfamily of our approach, Convolutional Neural Networks (CNNs).
A \textit{neural network} is a composition of $L$ parametric functions referred to as layers, where each layer is considered a representation of the input domain~\citep{Goodfellow16}. One layer $l_i$, such as $i \in \{1, ...,L\}$, contains neurons, which are small units that compute one element of the layer's output. The~layer $l_i$ takes as input the output of its previous layer $l_{i-1}$ and applies a transformation to compute its own output. The~behavior of these transformations is controlled by a set of parameters $\theta_i$ for each layer and an activation sublayer to shape the nonlinearity of the network. These parameters are called weights and link the input of the previous layer to the output of the current layer based on matrix multiplication. This process is also referred to as feedforward propagation in the deep learning literature and is the constituent of multilayer perceptrons (MLPs). A~neural network is usually called ``deep'' when it contains more than one layer between its input and output layer.
Following the good performance of CNN architectures in image recognition~\citep{Huang17} and natural language processing~\citep{Sutskever14,Devlin19}, CNNs have started to be adopted for time series analysis~\citep{Cristian17}. 
\textit{CNNs} are neural networks that use convolution in place of general matrix multiplication in at least one of their layers~\citep{Goodfellow16}. A~convolution can be seen as applying and sliding a filter over the time series. The~use of different types, numbers and sequences of filters allow the learning of multiple discriminative features (feature maps) useful for the classification~task.

\subsection{MTS~Classifiers}
\label{rw_mts_classifiers}
The state-of-the-art MTS classifiers are usually grouped into three categories: sim- ilarity-based, feature-based and deep learning~methods.

Similarity-based methods make use of similarity measures to compare two MTS (e.g., Euclidean distance). Dynamic Time Warping (DTW) has been shown to be the best similarity measure to use along with the k-Nearest Neighbors (k-NN)~\citep{Seto15}. 
DTW is not a distance metric as it does not fully satisfy the required properties (the triangle inequality in particular), but~its use as similarity measure along with the NN-rule is valid~\citep{Vidal85}.
There are two versions of kNN-DTW for MTS, dependent (DTW$_{D}$) and independent (DTW$_{I}$), and~neither dominates the other~\citep{Shokoohi17}.  DTW$_{I}$ measures the cumulative distances of all dimensions independently measured under DTW. DTW$_{D}$ uses a similar calculation with single-dimensional time series; it considers the squared Euclidean accumulated distance over the multiple~dimensions. 

Next, feature-based methods can be categorized into two families: shapelet-based and Bag-of-Words (BoW) classifiers. Shapelets models (gRSF~\citep{Karlsson16} and UFS~\citep{Wistuba15}) use subsequences (shapelets) to transform the original time series into a lower-dimensional space that is easier to classify. On~the other hand, BoW models (LPS~\citep{Baydogan16}, mv-ARF~\citep{Tuncel18}, SMTS~\citep{Baydogan14} and WEASEL+MUSE~\citep{Schafer17}) convert time series into a bag of discrete words and~use a histogram of words representation to perform the classification.
WEASEL+ MUSE shows better results compared to gRSF, LPS, mv-ARF, SMTS and UFS on average (20 MTS datasets). WEASEL+MUSE generates a BoW representation by applying various sliding windows with different sizes on each discretized dimension (Symbolic Fourier Approximation) to capture features (unigrams, bigrams and dimension identification). Following a feature selection with chi-square test, it classifies the MTS based on a logistic~regression.

{Finally, deep learning methods (FCN~\citep{Wang17}, MLSTM-FCN~\citep{Karim19}, MTEX-CNN~\citep{Assaf19}, \linebreak \mbox{ResNet~\citep{He16}}, TapNet~\citep{Zhang20} and TST~\citep{Zerveas21}) use Long-Short Term Memory (LSTM), Convolutional Neural Networks (CNN) or Transformers.} According to the results published and our experiments, the~current state-of-the-art model (MLSTM-FCN) is proposed in~\citep{Karim19} and consists of a LSTM layer and a stacked CNN layer along with squeeze-and-excitation blocks to generate latent features.
A recent network, TapNet~\citep{Zhang20}, also consists of a LSTM layer and a stacked CNN layer, followed by an attentional prototype network. However, TapNet shows lower accuracy results ({\url{https://github.com/xuczhang/xuczhang.github.io/blob/master/papers/aaai20_tapnet_full.pdf}} (accessed on 1 November 2021)) on average on the 30 public UEA MTS datasets compared to MLSTM-FCN (MLSTM-FCN results presented in Table~\ref{tab:UEA}).
There is no basis of comparison for MLSTM-FCN with MTEX-CNN~\citep{Assaf19} as MTEX-CNN has not been evaluated on public datasets. As~illustrated in Figure~\ref{fig:mtex}, MTEX-CNN is a two-stage CNN network which first extracts information relative to each feature with 2D convolution filters and then extracts information relative to time with 1D convolution filters. The~output feature map is fed into fully connected layers for~classification. 

Therefore, in~this work, we choose to benchmark XCM to the best-in-class for each similarity-based, feature-based and deep learning category (DTW$_{D}$/DTW$_{I}$, WEASEL+ MUSE and MLSTM-FCN classifiers). We also include MTEX-CNN in the benchmark to demonstrate the superiority of our approach as MTEX-CNN has not been evaluated on the public UEA~datasets.

\subsection{Explainability}
\label{sec:soa_explainability}
In addition to their prediction performance, machine learning methods have to be assessed on how they can support their decisions with explanations. 
Two levels of explanations are generally distinguished: global and local~\citep{Du20}. Global explainability means that explanations concern the overall behavior of the model across the full dataset, while local explainability informs the user about a particular prediction. 
As previously introduced with the example of the GDPR, our new CNN approach needs to be able to support each individual prediction. Thus, we present in this section the local explainability methods for~CNNs.

CNNs classifiers do not provide explainability-by-design at the local level. Thus, some post hoc model-agnostic explainability methods could be used. These methods provide explanations for any machine learning model. They treat the model as a black-box and do not inspect internal model parameters. The~main line of work consists of approximating the decision surface of a model using an explainable one (e.g., LIME~\citep{Ribeiro16}, SHAP~\citep{Lundberg17}, Anchors~\citep{Ribeiro18} and LORE~\citep{Guidotti19}). However, the~explanations from the surrogate models cannot be perfectly faithful with respect to the original model~\citep{Rudin19}, which is a prerequisite for numerous~applications.

Then, some post hoc model-specific explainability methods exist. These methods are specifically designed to extract explanations for a particular model. They usually derive explanations by examining internal model structures and parameters. The~approaches based on back-propagation are seen as the state-of-the-art explainability methods for deep learning models~\citep{Ancona18}. 
Methods based on back-propagation (e.g., Gradient Explanation~\citep{Erhan09}, Guided Backpropagation~\citep{Springenberg15}, $\varepsilon$-Layer-wise Relevance Propagation~\citep{Bach15}, Gradient $\odot$ Input~\citep{Shrikumar16}, Integrated Gradients~\citep{Sundararajan17}, DeepLift~\citep{Shrikumar17} and Grad-CAM~\citep{Selvaraju19}) calculate the gradient, or~its variants, of~a particular output with respect to the input using back-propagation to derive the contribution of features. 
In particular, Gradient-weighted Class Activation Mapping (Grad-CAM)~\citep{Selvaraju19} has proven to be an adequate method for supporting CNNs predictions. Grad-CAM identifies the regions of the input data that are important for predictions in CNNs using the class-specific gradient information. The~method has been shown to provide faithful explanations with regard to the model~\citep{Adebayo18}. The~faithfulness of the explanations provided by Grad-CAM is shown following a methodology based on model parameter and data randomization tests.
However, the~precision of the explanations provided by Grad-CAM, i.e.,~the fraction of explanations that are relevant to a prediction, can vary across CNN architectures as Grad-CAM is sensitive to the upsampling processes on feature maps to match the input data~dimensions.

Therefore, we support the predictions of our new CNN model XCM with Grad-CAM, a~post hoc model-specific explainability method which provides faithful explanations at local level. The~design of our network architecture avoids upsampling processes and~enables Grad-CAM to identify the observed variables and timestamps of the input data that are important for predictions more precisely as compared to what the current explainable deep learning MTS classifier MTEX-CNN~give. 

{Table~\ref{tab:MTSClassifiers} presents an overview of the challenges addressed by the state-of-the-art MTS classifiers and how we position our new method XCM. We evaluate the classification performance of XCM and its explainability in Section~\ref{XCM_results}.
The next section presents XCM in~details.}

\nointerlineskip
\begin{specialtable}[H]
\widetable
\caption{Overview of the state-of-the-art MTS~classifiers.}
\label{tab:MTSClassifiers}
\setlength{\cellWidtha}{\columnwidth/7-2\tabcolsep+0.9in}
\setlength{\cellWidthb}{\columnwidth/7-2\tabcolsep-0.3in}
\setlength{\cellWidthc}{\columnwidth/7-2\tabcolsep-0.3in}
\setlength{\cellWidthd}{\columnwidth/7-2\tabcolsep-0.17in}
\setlength{\cellWidthe}{\columnwidth/7-2\tabcolsep-0.17in}
\setlength{\cellWidthf}{\columnwidth/7-2\tabcolsep-0.05in}
\setlength{\cellWidthg}{\columnwidth/7-2\tabcolsep-0.17in}
\scalebox{1}[1]{\begin{tabularx}{.95\columnwidth}{>{\PreserveBackslash}m{\cellWidtha}>{\PreserveBackslash\centering}m{\cellWidthb}>{\PreserveBackslash\centering}m{\cellWidthc}>{\PreserveBackslash\centering}m{\cellWidthd}>{\PreserveBackslash\centering}m{\cellWidthe}>{\PreserveBackslash\centering}m{\cellWidthf}|>{\PreserveBackslash\centering}m{\cellWidthg}}
		\toprule
		&\textbf{ED}&\textbf{DTW}&\textbf{MLSTM FCN}&\textbf{MTEX CNN}&\textbf{WEASEL+ MUSE}&\textbf{XCM}\\
		\midrule
		\multicolumn{1}{l}{\textbf{Performance}} & & & & & &\\
		\hspace{.4cm}Small Datasets & & & & & \checkmark & \checkmark \\
		\hspace{.4cm}Large Datasets & & & \checkmark & & & \checkmark\\
		& & & & & & \\
		\multicolumn{1}{l}{\textbf{Explainability}} & & & & & &\\
		\hspace{.4cm}Faithful Explainability & \checkmark & \checkmark & & \checkmark & & \checkmark\\
		\bottomrule
\end{tabularx}}
\end{specialtable}

\section{XCM} \label{sec3}
In this section, we present our new eXplainable Convolutional neural network for Multivariate time series classification (XCM). The~first part details the architecture of the network, and the second part explains how XCM can provide explanations by identifying the observed variables and timestamps of the input data that are important for~predictions.

\subsection{Architecture}
\label{sec:XCM}
\begin{figure}[H]
	\centering
	\includegraphics[width=1\linewidth]{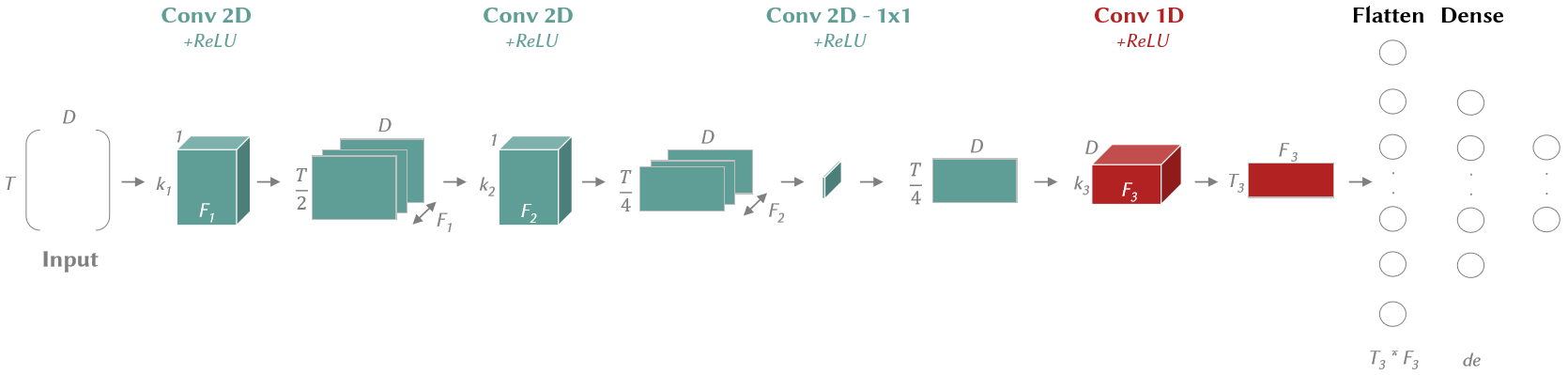}
	\caption{MTEX-CNN architecture. Abbreviations: \textit{D}---number of observed variables, \textit{de}---dense layer size, \textit{F}---number of filters, \textit{k}---kernel size and \textit{T}---time series~length.}
	\label{fig:mtex}
\end{figure}

\begin{figure}[H]
	\centering
	\includegraphics[width=1\linewidth]{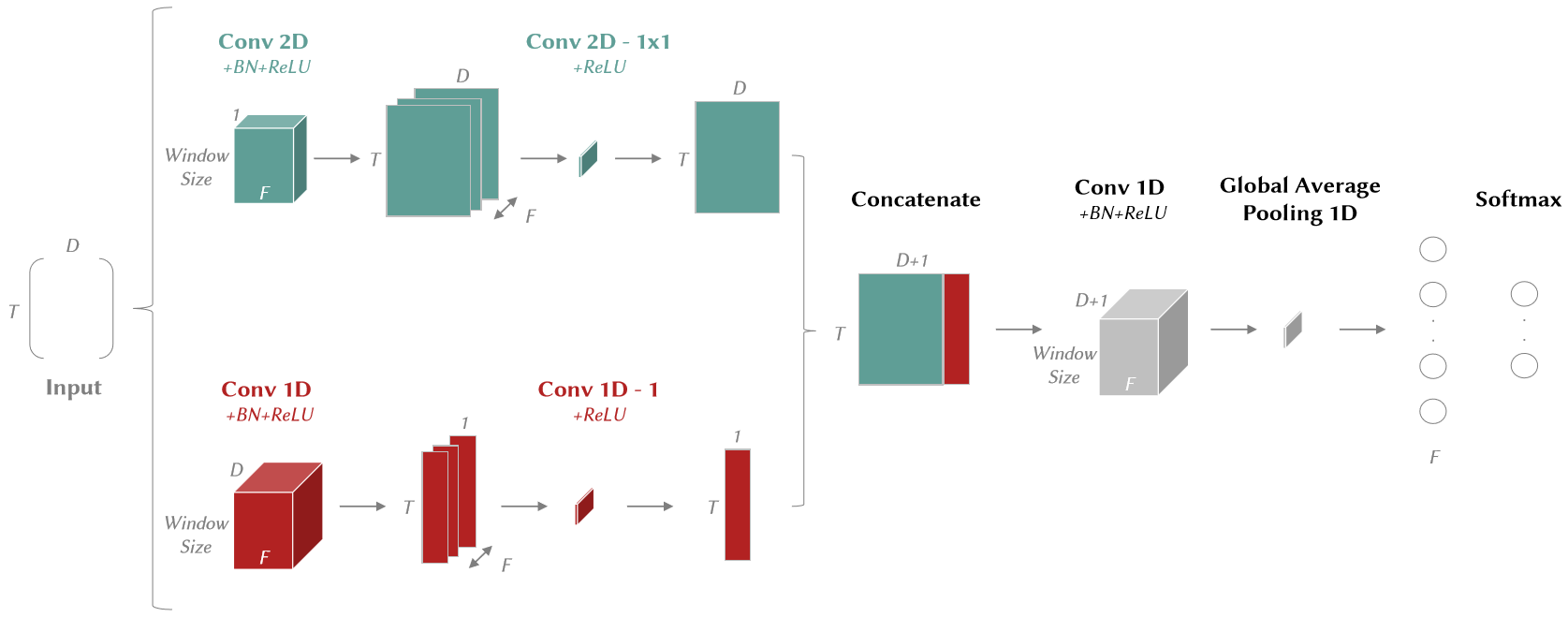}
	\caption{XCM architecture. Abbreviations: \textit{BN}---Batch Normalization, \textit{D}---number of observed variables, \textit{F}---number of filters, \textit{T}---time series length and \textit{Window Size}---kernel size, which corresponds to the time window~size.}
	\label{fig:xcm}
\end{figure}

Our approach aims to design a new compact and explainable CNN architecture that performs well on both the large and small UEA datasets.
As illustrated in \mbox{Figure~\ref{fig:mtex}}, a~recent explainable CNN, MTEX-CNN~\citep{Assaf19}, proposes to use 2D and 1D convolution filters in sequence to extract key MTS information, i.e.,~information relative to the observed variables and time, respectively.
However, CNN architectures such as MTEX-CNN have significant limitations. 
The use of 2D and 1D convolution filters in sequence means that the features related to time (features maps from 1D convolution filters) are extracted from the processed features related to observed variables (features maps from 2D convolution filters). Therefore, features related to time cannot fully incorporate the timing information from the input data and~can only partially reflect the necessary information to discriminate between the different classes. 
Thus, \textit{(i)} our approach XCM extracts both features related to observed variables (2D convolution filters) and time (1D convolution filters) directly from the input data, which leads to more discriminative features by incorporating all the relevant information and ultimately to a better classification performance on average than the 2D/1D sequential approach (see results in Section~\ref{sec:performance}).
Then, a~CNN architecture using fully connected layers to perform classification, especially with the size of the first layer depending on the time series length as in MTEX-CNN, is prone to overfitting and can lead to the explosion of the number of trainable parameters.
Thus, \textit{(ii)} the output feature maps of XCM are processed with a 1D global average pooling before being input to a softmax layer for classification. The~use of 1D global average pooling followed by a softmax layer for classification reduces the number of parameters and improves the generalization ability of the network compared to fully connected layers. Global average pooling consists of summarizing each feature map by its average. This operation improves the generalization ability of the network, as it does not have parameters to train, and it provides robustness to spatial translations of the input~\citep{Lin14}. In~the possible cases when the sequences of events in an MTS change, the~robustness to spatial translation ensures that the classification result is not modified.
Finally, the~use of non fully padded convolution filters as in MTEX-CNN can lead to an imprecise identification of the regions of the input data that are important for predictions as Grad-CAM is sensitive to upsampling processes. Therefore, \textit{(iii)} the 2D and 1D convolution filters of XCM are fully padded. As~detailed in the next section, the~output feature maps can then be analyzed with the Grad-CAM explainability method without altering the precision of the explanations through upsampling processes. Figure~\ref{fig:xcm} illustrates XCM, and the following paragraphs detail the~architecture.

\unskip
\vspace{-6pt}
\unskip

Firstly, XCM extracts information relative to the observed variables with 2D convolution filters (upper green part in Figure~\ref{fig:xcm}). 
This upper part is composed of one 2D convolutional block which is then converted to one feature map to reduce the number of parameters with a $1 \times 1$ convolution filter.
The convolutional block contains a 2D convolution layer followed by a batch normalization layer~\citep{Ioffe15} and a ReLU activation layer~\citep{Nair10}.
We set the kernel size of the 2D convolution filters to $Window \, Size \times 1$, where $Window \, Size$ is a hyperparameter which specifies the time window size, i.e.,~the size of the subsequence of the MTS expected to be interesting to extract discriminative features, and~$\times 1$ means for each observed variable.
Thus, these 2D convolution filters (number: $F$ in Figure~\ref{fig:xcm}) allow the extraction of features per observed variable. The~features are extracted using a sliding window (strides equal to 1), and we use padding instead of half padding to keep the dimension of the feature maps the same as the input data. The~padding allows us to avoid using upsampling and interpolation methods on the features maps when building the \textit{attribution maps}, i.e.,~the heatmaps of dimensions $T \times D$ that identify the regions of the input data that are important for predictions (detailed in the next section). Then, batch normalization brings normalization at the layer level, and it enables faster convergence and better generalization of the network~\citep{Bjorck18}. In addition, the~ReLU activation layer induces nonlinearity in the network. Next, the~output feature maps are fed into a module ($1 \times 1$ convolution filter)~\citep{Szegedy15} which reduces the number of parameters. It projects the feature maps into one following a channel-wise~pooling. 

In parallel, XCM extracts information relative to time with 1D convolution filters (lower red part in Figure~\ref{fig:xcm}). This lower part is the same as the upper part, except~that the 2D convolution filters are replaced by 1D. We set the kernel size of the 1D convolution filters to $Window \, Size \times D$, where $Window \, Size$ is the same hyperparameter as 2D convolution filters and $D$ is the number of observed variables of the input data.
The 1D convolution filters slide over the time axis only (stride equals to 1) and capture the interaction between the different time series.
Following the use of padding, the~output feature map of this lower part has a dimension of $T \times 1$, with~$T$ the time series length of the input data.
The use of padding, similar to 2D convolution filters, allows us to avoid using upsampling of the features maps on the dimension related to the information extracted (time-$T$) when building the \textit{attribution maps} (detailed in the next section).

In the following step, the~output feature maps from these two parts are concatenated and form a feature map of dimensions $T \times (D+1)$.
We apply the same 1D convolution block (1D convolution layer---$F$ filters, kernel size $Window \, Size \times (D+1)$, stride 1 and padding + batch normalization + ReLU activation layer) as presented in the previous paragraph to slide over the time axis and capture the interaction between the features extracted.
Finally, we add a 1D global average pooling on the output feature maps and perform classification with a softmax layer. As~previously introduced, the~use of global average pooling instead of fully connected layers improves the generalization ability of the~network.

In order to assess the potential advantage of concatenating the 2D and 1D convolution blocks instead of having them in sequence, independently from the choice of the classification layers (fully connected layers as in MTEX-CNN versus 1D global average pooling with a softmax layer in XCM), we include in our experiments in Section~\ref{sec:performance} a variant of XCM (XCM-Seq). XCM-seq is the same as XCM except that the 2D and 1D convolution blocks are in sequence. The~next section presents how the architecture of XCM allows the communication of explanations supporting the model predictions with~Grad-CAM.

\subsection{Explainability}
The new CNN architecture of XCM has been designed to enable the precise identification of the observed variables and timestamps that are important for predictions based on Gradient-weighted Class Activation Mapping (Grad-CAM)~\citep{Selvaraju19}.
As presented in \mbox{Section~\ref{sec:soa_explainability}}, Grad-CAM identifies the regions of the input data that are important for predictions in CNNs using the class-specific gradient information. More specifically, Grad-CAM can output two types of attribution maps from XCM architecture: one related to observed variables and another one related to time. Attribution maps are heatmaps of the same size as the input data where some colors indicate features that contribute positively to the activation of the target output~\citep{Ancona18}. These attribution maps constitute the explanations provided to support XCM model predictions and are available at the sample level. The~following paragraphs explain how we adapt Grad-CAM for~XCM.

In order to build the first attribution map related to observed variables, Grad-CAM is applied to the output feature maps of the 2D convolution layer which uses convolution filters per observed variable (first block in the upper green part in Figure~\ref{fig:xcm}).
To obtain the class-discriminative attribution map, $L_{2D}^c \in \mathbb{R}^{T \times D}$ with $T$ the time series length and $D$ the number of observed variables, we first compute the gradient of the score for class $c$ ($y_c$) with respect to feature map activations $A^k$ of the convolutional layer, i.e.,~$\frac{\partial y_c}{\partial A^k}$ with $k \in [1,\dots,F]$ the identifier of the feature map. These gradients flowing back are global-average-pooled over the time series length ($T$) and observed variables ($D$) dimensions (indexed by i and j, respectively) to obtain the weight of each feature map. Thus, as~regards the feature map $k$, we calculate the weight as:
\begin{equation}	
	{w_{k}^{c} = \frac{1}{T \times D} \sum_{i} \sum_{j} \frac{\partial y_c}{\partial A_{ij}^{k}}}
\end{equation}

We then use the weights to compute a weighted combination between all the feature maps for that particular class and~use a ReLU to keep only the positive attributions to the predictions (Equation (3)).
\begin{equation}
	{L_{2D}^c = ReLU \left(\sum_{k} w_{k}^c \, A^{k}\right)}
\end{equation}

The second attribution map, $L_{1D}^c$, relates to time and is built on the same principle. Grad-CAM is applied to the output feature maps of the 1D convolution layer which uses convolution filters sliding over the time axis (first block in the lower red part in Figure~\ref{fig:xcm}). With~respect to the feature maps activations $M$ and the class $c$, we calculate $L_{1D}^c$ as:
\begin{equation}	
	{q_{k}^{c} = \frac{1}{T} \sum_{i} \frac{\partial y_c}{\partial M_{i}^{k}}}
\end{equation}
\begin{equation}
	{L_{1D}^c = ReLU \left(\sum_{k} q_{k}^c \, M^{k}\right)}
\end{equation}

Thus, $L_{1D}^c$ has $T \times 1$ as dimensions. We then upsample it to match the input data dimensions $T \times D$ with a bilinear interpolation in order to obtain the attribution map. This operation does not alter the time attribution results as the padding on the 1D convolution filters ensured that the feature extraction over the time dimension has kept the time series length. Therefore, the~upsampling only replicates the results over the observed variables. An example of observed variables and time attribution maps on a synthetic dataset is presented in Section~\ref{sec:res_explainability}.

Before discussing the performance and explainability results of XCM, we present in the next section the evaluation~setting.

\section{Evaluation} \label{sec4}
In this section, we present the methodology employed (datasets, algorithms, hyperparameters and metrics) to evaluate our~approach.

\subsection{Datasets}
We benchmarked XCM on the 30 currently available public UEA MTS datasets~\citep{Bagnall18}. We kept the train/test splits provided in the archive. The~characteristics of each dataset are presented in Table~\ref{tab:Datasets}.

\subsection{Algorithms}
\label{sec:eval_algo}
We compare our algorithm XCM implemented in Python 3.6 (code available on GitHub \url{https://github.com/XAIseries/XCM}) to the state-of-the-art MTS classifiers, as~detailed in Section~\ref{rw_mts_classifiers}, and~to the variant XCM-Seq: 

\begin{itemize}
	\item DTW$_{D}$, DTW$_{I}$ and ED---with and without normalization (n): we reported the results published in the UEA archive~\citep{Bagnall18};
	\item MLSTM-FCN: we used the implementation available {(\url{https://github.com/houshd/MLSTM-FCN} (accessed on 1 November 2021))} and ran it with the parameter settings recommended by the authors in the paper~\citep{Karim19} (128-256-128 filters, kernel sizes 8/5/3, initialization of convolution kernels Uniform He, reduction ratio of 16, 250 training epochs, dropout of 0.8, Adam optimizer) and with the following hyperparameters: batch size $\{1, 8, 32\}$, number of LSTM cells $\{8,64,128\}$;
	\item MTEX-CNN: we implemented the algorithm with Keras in Python 3.6 based on the description of the paper~\citep{Assaf19}. We ran it with the parameter settings recommended by the authors (Stage 1: two convolution layers with half padding and ReLU activation, kernel sizes $8 \times 1$ and $6 \times 1$, strides $2 \times 1$, feature maps 64 and 128, dropout 0.4. Stage 2: one convolution layer with ReLU activation, strides 2, kernel size 2, feature maps 128, dropout 0.4. Dense layer dimension 128 and L2 regularization 0.2) and with the following hyperparameter: batch size $\{1, 8, 32\}$;
	\item \textls[-30]{WEASEL+MUSE: we used the implementation available} {(\url{https://github.com/patrickzib/SFA} (accessed on 1 November 2021))} and ran it with the parameter settings recommended by the authors in the paper~\citep{Schafer17} (chi = 2, bias = 1, p = 0.1, c = 5 and L2R\_LR\_DUAL solver) and with the following hyperparameters: SFA word lengths $\{2,4,6\}$, SFA quantization method \{equi-depth, equi-frequency\}, windows length [4, max(MTS length)];
	\item XCM: we implemented the algorithm with Keras in Python 3.6. 2D convolution layers with: 128 feature maps, kernel size: $Window \, Size \times 1$, strides $1 \times 1$, padding \textit{same} and ReLU activation. In addition,~1D convolution layers with: 128 feature maps, kernel size: $Window \, Size$, strides 1, padding \textit{same} and ReLU activation. The~hyperparameters are: batch size $\{1, 8, 32\}$ and $Window \, Size$ (the time window size---kernel size), expressed as a percentage of the total size of the MTS $[20\%,40\%,60\%,80\%,100\%]$;
	\item XCM-Seq: XCM variant with 2D and 1D convolution blocks in sequence (see description in Section~\ref{sec:XCM}). We used the same setting as XCM.
\end{itemize}

All the networks that we implemented (XCM, XCM-Seq and MTEX-CNN) were trained with 100 epochs, the~categorical crossentropy loss and the Adam optimization (computing infrastructure: Debian 8 operating system, GPU NVIDIA GeForce RTX 2080 Ti with 11Gb GRAM and 96Gb of RAM).
Concerning Grad-CAM, we used the implementation available for Keras ({\url{https://github.com/jacobgil/keras-grad-cam}} (accessed on 1 November 2021)).

\end{paracol}
\begin{specialtable}[H]
\widetable
\caption{UEA MTS datasets. Abbreviations: AS---Audio Spectra, ECG---Electrocardiogram, EEG---Electroencephalogram, HAR---Human Activity Recognition and MEG---Magnetoencephalography.}
\label{tab:Datasets}
\setlength{\cellWidtha}{\columnwidth/7-2\tabcolsep+1.1in}
\setlength{\cellWidthb}{\columnwidth/7-2\tabcolsep-0.1in}
\setlength{\cellWidthc}{\columnwidth/7-2\tabcolsep-0.2in}
\setlength{\cellWidthd}{\columnwidth/7-2\tabcolsep-0.2in}
\setlength{\cellWidthe}{\columnwidth/7-2\tabcolsep-0.2in}
\setlength{\cellWidthf}{\columnwidth/7-2\tabcolsep-0.2in}
\setlength{\cellWidthg}{\columnwidth/7-2\tabcolsep-0.2in}
\scalebox{1}[1]{\begin{tabularx}{\columnwidth}{>{\PreserveBackslash\raggedright}m{\cellWidtha}>{\PreserveBackslash\centering}m{\cellWidthb}>{\PreserveBackslash\centering}m{\cellWidthc}>{\PreserveBackslash\centering}m{\cellWidthd}>{\PreserveBackslash\centering}m{\cellWidthe}>{\PreserveBackslash\centering}m{\cellWidthf}>{\PreserveBackslash\centering}m{\cellWidthg}}
		\toprule
		\textbf{Datasets} & \textbf{Type} & \textbf{Train} & \textbf{Test} & \textbf{Length} & \textbf{Dimensions} & \textbf{Classes}\\
		\midrule
		Articulary Word Recognition & Motion & 275 & 300 & 144 & 9 & 25\\
		Atrial Fibrilation & ECG & 15 & 15 & 640 & 2 & 3\\
		Basic Motions & HAR & 40 & 40 & 100 & 6 & 4\\
		Character Trajectories & Motion & 1422 & 1436 & 182 & 3 & 20\\
		Cricket & HAR & 108 & 72 & 1197 & 6 & 12\\
		Duck Duck Geese & AS & 60 & 40 & 270 & 1345 & 5\\
		Eigen Worms & Motion & 128 & 131 & 17,984 & 6 & 5\\
		Epilepsy & HAR & 137 & 138 & 206 & 3 & 4\\
		Ering & HAR & 30 & 30 & 65 & 4 & 6\\
		Ethanol Concentration & Other & 261 & 263 & 1751 & 3 & 4\\
		Face Detection & EEG/MEG & 5890 & 3524 & 62 & 144 & 2\\
		Finger Movements & EEG/MEG & 316 & 100 & 50 & 28 & 2\\
		Hand Movement Direction & EEG/MEG & 320 & 147 & 400 & 10 & 4\\
		Handwriting & HAR & 150	& 850 & 152 & 3 & 26\\ 
		Heartbeat & AS & 204 & 205 & 405 & 61 & 2\\
		Insect Wingbeat & AS & 30,000 & 20,000 & 200 & 30 & 10\\
		Japanese Vowels & AS & 270 & 370 & 29 & 12 & 9\\
		Libras & HAR & 180 & 180 & 45 & 2 & 15\\
		LSST & Other & 2459 & 2466 & 36 & 6 & 14\\
		Motor Imagery & EEG/MEG & 278 & 100 & 3000 & 64 & 2\\
		NATOPS & HAR & 180 & 180 & 51 & 24 & 6\\
		PenDigits & Motion & 7494 & 3498 & 8 & 2 & 10\\
		PEMS-SF & Other & 267 & 173 & 144 & 963 & 7\\
		Phoneme & AS & 3315 & 3353 & 217 & 11 & 39\\
		Racket Sports & HAR & 151 & 152 & 30 & 6 & 4\\
		Self Regulation SCP1 & EEG/MEG & 268 & 293 & 896 & 6 & 2\\
		Self Regulation SCP2 & EEG/MEG & 200 & 180 & 1152 & 7 & 2\\
		Spoken Arabic Digits & AS & 6599 & 2199 & 93 & 13 & 10\\
		Stand Walk Jump & ECG & 12 & 15 & 2500 & 4 & 3\\
		U Wave Gesture Library & HAR & 120 & 320 & 315 & 3 & 8\\
		\bottomrule
\end{tabularx}}
\end{specialtable}
\begin{paracol}{2}
\switchcolumn
\vspace{-12pt}

\subsection{Hyperparameters}
\label{sec:hyperparameters}
For each dataset, hyperparameters were set by grid search based on the best average accuracy following a stratified 5-fold cross-validation on the training~set.

\subsection{Metrics}
{For each dataset, we computed the classification accuracy---the metric used to benchmark the MTS classifiers on the public UEA datasets~\citep{Bagnall18}}. Then, we presented the average rank and the number of wins/ties to compare the different classifiers on the same datasets. Finally, we presented the critical difference diagram~\citep{Demsar06}, the~statistical comparison of multiple classifiers on multiple datasets based on the nonparametric Friedman test, to~show the overall performance of XCM. We used the implementation available in R package scmamp ({\url{https://www.rdocumentation.org/packages/scmamp/versions/0.2.55/topics/plotCD}} (accessed on 1 November 2021)).

\end{paracol}
\begin{specialtable}[H]
\widetable
\caption{Accuracy results on the UEA MTS datasets. Abbreviations: Batch---Batch Size, DW$_{D}$---DTW$_{D}$, DW$_{I}$---DTW$_{I}$, MC---MTEX-CNN, MF---MLSTM-FCN, Win \%---Time Window Size, WM---WEASEL+MUSE and XC---XCM.}
\label{tab:UEA}
\setlength{\cellWidtha}{\columnwidth/14-2\tabcolsep+1.3in}
\setlength{\cellWidthb}{\columnwidth/14-2\tabcolsep-0.13in}
\setlength{\cellWidthc}{\columnwidth/14-2\tabcolsep-0.13in}
\setlength{\cellWidthd}{\columnwidth/14-2\tabcolsep-0.13in}
\setlength{\cellWidthe}{\columnwidth/14-2\tabcolsep-0.13in}
\setlength{\cellWidthf}{\columnwidth/14-2\tabcolsep-0.13in}
\setlength{\cellWidthg}{\columnwidth/14-2\tabcolsep-0.13in}
\setlength{\cellWidthh}{\columnwidth/14-2\tabcolsep-0.13in}
\setlength{\cellWidthi}{\columnwidth/14-2\tabcolsep-0.13in}
\setlength{\cellWidthj}{\columnwidth/14-2\tabcolsep-0.13in}
\setlength{\cellWidthk}{\columnwidth/14-2\tabcolsep-0.13in}
\setlength{\cellWidthl}{\columnwidth/14-2\tabcolsep-0.11in}
\setlength{\cellWidthm}{\columnwidth/14-2\tabcolsep-0.13in}
\setlength{\cellWidthn}{\columnwidth/14-2\tabcolsep+0.21in}
\scalebox{1}[1]{\begin{tabularx}{\columnwidth}{>{\PreserveBackslash\raggedright}m{\cellWidtha}>{\PreserveBackslash\raggedright}m{\cellWidthb}>{\PreserveBackslash\centering}m{\cellWidthc}>{\PreserveBackslash\centering}m{\cellWidthd}>{\PreserveBackslash\centering}m{\cellWidthe}>{\PreserveBackslash\centering}m{\cellWidthf}>{\PreserveBackslash\centering}m{\cellWidthg}>{\PreserveBackslash\centering}m{\cellWidthh}>{\PreserveBackslash\centering}m{\cellWidthi}>{\PreserveBackslash\centering}m{\cellWidthj}>{\PreserveBackslash\centering}m{\cellWidthk}>{\PreserveBackslash\centering}m{\cellWidthl}|>{\PreserveBackslash\centering}m{\cellWidthm}>{\PreserveBackslash\centering}m{\cellWidthn}}
		\toprule
		\multirow{2}{*}{\textbf{Datasets}} & \multirow{2}{.4cm}{\centering\textbf{XC}} & \multirow{2}{.5cm}{\centering\textbf{XC Seq}} & \multirow{2}{.8cm}{\centering\textbf{MC}} & \multirow{2}{.8cm}{\centering\textbf{MF}} & \multirow{2}{.8cm}{\centering\textbf{WM}} & \multirow{2}{.3cm}{\centering\textbf{ED}} & \multirow{2}{.5cm}{\centering\textbf{DW$_{I}$}} & \multirow{2}{.5cm}{\centering\textbf{DW$_{D}$}} & \multirow{2}{.6cm}{\centering\textbf{ED (n)}} & \multirow{2}{.6cm}{\centering\textbf{DW$_{I}$ (n)}} & \multirow{2}{.6cm}{\centering\textbf{DW$_{D}$ (n)}} & \multicolumn{2}{p{2.2cm}}{\centering\textbf{XC Parameters}} \\
		& & & & & & & & & & & & \textbf{Batch} & \textbf{Win} \%\\
		\midrule
		Articulary Word Recognition & 98.3 & 92.7 & 92.3 & 98.6 & \textbf{99.3} & 97.0 & 98.0 & 98.7 & 97.0 & 98.0 & 98.7 & 32 & 80\\
		Atrial Fibrilation & \textbf{46.7} & 33.3 & 33.3 & 20.0 & 26.7 & 26.7 & 26.7 & 20.0 & 26.7 & 26.7 & 22.0 & 1 & 60\\
		Basic Motions & \textbf{100.0} & \textbf{100.0} & \textbf{100.0} & \textbf{100.0} & \textbf{100.0} & 67.5 & \textbf{100.0} & 97.5 & 67.6 & \textbf{100.0} & 97.5 & 32 & 20\\
		Character Trajectories & \textbf{99.5} & 98.8 & 97.4 & 99.3 & 99.0 & 96.4 & 96.9 & 99.0 & 96.4 & 96.9 & 98.9 & 32 & 80\\
		Cricket & \textbf{100.0} & 93.1 & 90.3 & 98.6 & 98.6 & 94.4 & 98.6 & \textbf{100.0} & 94.4 & 98.6 & \textbf{100.0} & 32 & 20\\
		Duck Duck Geese & \textbf{70.0} & 52.5 & 65.0 & 67.5 & 57.5 & 27.5 & 55.0 & 60.0 & 27.5 & 55.0 & 60.0 & 8 & 80\\
		Eigen Worms & 43.5 & 45.0 & 41.9 & 80.9 & \textbf{89.0} & 55.0 & 60.3 & 61.8 & 54.9 &  & 61.8 & 32 & 40\\
		Epilepsy & \textbf{99.3} & 93.5 & 94.9 & 96.4 & \textbf{99.3} & 66.7 & 97.8 & 96.4 & 66.6 & 97.8 & 96.4 & 32 & 20\\
		Ering & \textbf{13.3} & \textbf{13.3} & \textbf{13.3} & \textbf{13.3} & \textbf{13.3} & \textbf{13.3} & \textbf{13.3} & \textbf{13.3} & \textbf{13.3} & \textbf{13.3} & \textbf{13.3} & 32 & 20\\
		Ethanol Concentration & \textbf{34.6} & 31.6 & 30.8 & 29.4 & 31.6 & 29.3 & 30.4 & 32.3 & 29.3 & 30.4 & 32.3 & 32 & 80\\
		Face Detection & \textbf{63.9} & 63.8 & 50.0 & 57.4 & 54.5 & 51.9 & 51.3 & 52.9 & 51.9 &  & 52.9 & 32 & 60\\
		Finger Movements & 60.0 & 60.0 & 49.0 & \textbf{61.0} & 54.0 & 55.0 & 52.0 & 53.0 & 55.0 & 52.0 & 53.0 & 32 & 40\\
		Hand Movement Direction & \textbf{44.6} & 40.1 & 18.9 & 37.8 & 37.8 & 27.9 & 30.6 & 23.1 & 27.8 & 30.6 & 23.1 & 32 & 80\\
		Handwriting & 41.2 & 38.6 & 24.6 & 54.9 & 53.1 & 37.1 & 50.9 & \textbf{60.7} & 20.0 & 31.6 & 28.6 & 32 & 60\\ 
		Heartbeat & \textbf{77.6} & 74.1 & 72.2 & 71.4 & 72.7 & 62.0 & 65.9 & 71.7 & 61.9 & 65.8 & 71.7 & 32 & 80\\
		Insect Wingbeat & 10.5 & 10.5 & 10.5 & 10.5 &  & \textbf{12.8} &  & 11.5 & \textbf{12.8} & & & 32 & 20\\
		Japanese Vowels & 98.6 & 94.6 & 95.1 & \textbf{99.2} & 97.8 & 92.4 & 95.9 & 94.9 & 92.4 & 95.9 & 94.9 & 32 & 80\\
		Libras & 84.4 & 79.4 & 81.1 & \textbf{92.2} & 89.4 & 83.3 & 89.4 & 87.2 & 83.3 & 89.4 & 87.0 & 32 & 80\\
		LSST & 61.2 & 54.2 & 31.5 & \textbf{64.6} & 62.8 & 45.6 & 57.5 & 55.1 & 45.6 & 57.5 & 55.1 & 32 & 100\\
		Motor Imagery & \textbf{54.0} & 53.0 & 50.0 & 53.0 & 50.0 & 51.0 & 39.0 & 50.0 & 51.0 &  & 50.0 & 8 & 40\\
		NATOPS & \textbf{97.8} & 93.9 & 88.3 & 96.7 & 88.3 & 85.0 & 85.0 & 88.3 & 85.0 & 85.0 & 88.3 & 32 & 40\\
		PenDigits & \textbf{99.1} & 96.7 & 87.8 & 99.0 & 96.9 & 97.3 & 93.9 & 97.7 & 97.3 & 93.9 & 97.7 & 8 & 60\\
		PEMS-SF & 75.7 & \textbf{80.9} & 11.6 & 69.9 &  & 70.5 & 73.4 & 71.1 & 70.5 & 73.4 & 71.1 & 32 & 80\\
		Phoneme & 22.5 & 11.9 & 2.6  & \textbf{27.5} & 19.0 & 10.4 & 15.1 & 15.1 & 10.4 & 15.1 & 15.1 & 32 & 40\\
		Racket Sports & 89.5 & 86.8 & 82.9 & 89.4 & \textbf{91.4} & 86.4 & 84.2 & 80.3 & 86.8 & 84.2 & 80.3 & 32 & 80\\
		Self Regulation SCP1 & \textbf{87.8} & 81.6 & 78.5 & 86.7 & 74.4 & 77.1 & 76.5 & 77.5 & 77.1 & 76.5 & 77.5 & 32 & 80\\
		Self Regulation SCP2 & 54.4 & \textbf{55.0} & 50.0 & 52.2 & 52.2 & 48.3 & 53.3 & 53.9 & 48.3 & 53.3 & 53.9 & 32 & 80\\
		Spoken Arabic Digits & \textbf{99.5} & 99.4 & 98.6 & 99.4 & 98.2 & 96.7 & 96.0 & 96.3 & 96.7 & 95.9 & 96.3 & 32 & 80\\
		Stand Walk Jump & 40.0 & 46.7 & \textbf{53.3} & 46.7 & 33.3 & 20.0 & 33.3 & 20.0 & 20.0 & 33.3 & 20.0 & 32 & 60\\
		U Wave Gesture Library & 89.4 & 81.9 & 81.2 & 86.3 & \textbf{90.3} & 88.1 & 86.9 & \textbf{90.3} & 88.1 & 86.8 & \textbf{90.3} & 32 & 100\\
		\midrule
		Average Rank & 2.3 & 5.0 & 7.2 & 3.5 & 4.0 & 7.1 & 5.9 & 4.8 & 7.4 & 6.4 & 5.3 & &\\
		Wins/Ties & 16 & 4 & 3 & 7 & 7 & 2 & 2 & 4 & 2 & 2 & 3 & &\\
		\bottomrule
\end{tabularx}}
\end{specialtable}
\begin{paracol}{2}
\switchcolumn

\section{Results} 
\label{XCM_results}
In this section, we first present the performance results of XCM on the public UEA datasets. Then, we illustrate how XCM can reconcile performance and explainability on a synthetic dataset. Finally, we end this section by showing that XCM outperforms the current most accurate state-of-the-art algorithm in a real-world application while providing faithful and more informative~explanations.

\subsection{Performance}
\label{sec:performance}
The accuracy results on the public UEA test sets of XCM and the other MTS classifiers are presented in Table~\ref{tab:UEA}. A~blank in the table indicates that the approach ran out of memory. The~best accuracy for each dataset is denoted in~boldface.

Firstly, we observe that XCM obtains the best average rank and the lowest rank variability across the datasets (rank: 2.3, standard error: 0.4), followed by MLSTM-FCN in second position (rank: 3.5, standard error: 0.5) and WEASEL+MUSE in third position (rank: 4.0, standard error: 0.5). Using the categorization of the datasets published in the archive website ({\url{http://www.timeseriesclassification.com/dataset.php}} (accessed on 1 November 2021)), we do not see any influence from the different train set sizes, MTS lengths, number of dimensions, number of classes and dataset types on XCM performance relative to the other classifiers on the UEA~datasets. 

More specifically, XCM exhibits better performance than MLSTM-FCN and WEASEL +MUSE on both the large (rank: 1.9, MLSTM-FCN rank: 2.1, WEASEL+MUSE rank: 4.6-train size $\geq$ 500, 23\% of the datasets) and small datasets (rank: 2.4, MLSTM-FCN rank: 4.0, WEASEL+MUSE rank: 3.9-train size $<$ 500, 77\% of the datasets).
We can assume that the more compact architecture of XCM compared to the other deep learning classifiers provides a better generalization ability on the UEA datasets (average rank on the number of trainable parameters: XCM 1.7, MLSTM-FCN: 1.9, MTEX-CNN: 2.0).
Furthermore, the~results confirm the superiority of the XCM approach based on the extraction in parallel and directly from the input data of features relative to the observed variables and time compared to the sequential approaches.
XCM outperforms both XCM-Seq and MTEX-CNN on average on the UEA datasets (rank: 2.3, XCM-Seq: 5.0, MTEX-CNN: 7.2).

With regard to the hyperparameter $Window \, Size$ of XCM, Figure~\ref{fig:time_windows} shows the average relative drop in performance across the datasets when using the other time window sizes than the one used in the best configuration given in Table~\ref{tab:UEA}. 
In order to evaluate the relative impact with respect to the range of performance, we defined four categories of datasets: datasets with XCM original accuracy$<$ 50\%, datasets with 50\% $\leq$ accuracy $<$ 75\%, datasets with 75\% $\leq$ accuracy $<$ 90\% and datasets with accuracy $\geq$ 90\%.
First, as~expected, we observe that the average relative impact of using suboptimal time window sizes is higher when XCM level of performance is low (average relative drop in accuracy: 13.1\% when XCM accuracy $<$ 50\% versus 3.0\% when XCM accuracy $\geq$ 90\%).
Then, the~average relative drop in accuracy when using suboptimal time window sizes is not negligible but remains limited in all the cases. This drop is below 15\% on average on the category where XCM has the lowest level of accuracy (13.1\% $\pm$ 3.2\%) and below 10\% on average across all the datasets (7.0\% $\pm$ 1.3\%).
\begin{figure}[H]
	\centering
	\includegraphics[width=.7\linewidth]{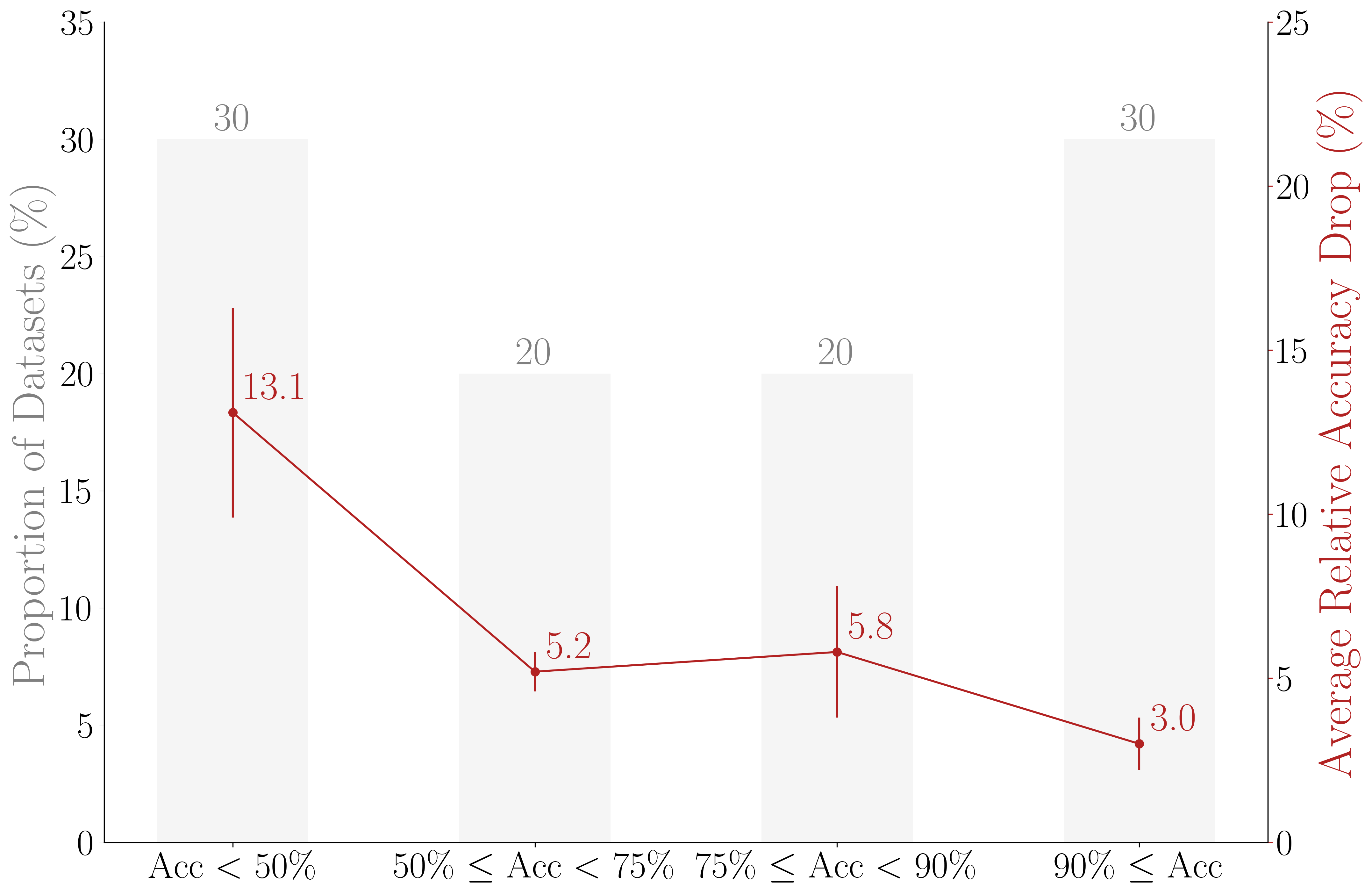}
	\caption{XCM average relative accuracy drop across the UEA datasets when using other time window sizes than the one used in the best configuration given in Table~\ref{tab:UEA}. The~performance drop is presented across four categories of datasets, defined according to XCM levels of accuracy shown in Table~\ref{tab:UEA}. Abbreviation: Acc---Accuracy.}
	\label{fig:time_windows}
\end{figure}
Finally, we performed a statistical test to evaluate the performance of XCM compared to the other MTS classifiers. We present in Figure~\ref{fig:CDPlot_MTS} the critical difference plot with alpha equals to 0.05 from results shown in Table~\ref{tab:UEA}. The~values correspond to the average rank, and the classifiers linked by a bar do not have a statistically significant difference. The~plot confirms the top three ranking as presented before (XCM: 1, MLSTM-FCN: 2, and WEASEL+MUSE: 3), without~showing a statistically significant difference between each other. We notice that XCM is the only classifier with a significant performance difference compared to DTW$_{D}$.

\begin{figure}[H]
	\centering
	\includegraphics[width=.9\linewidth]{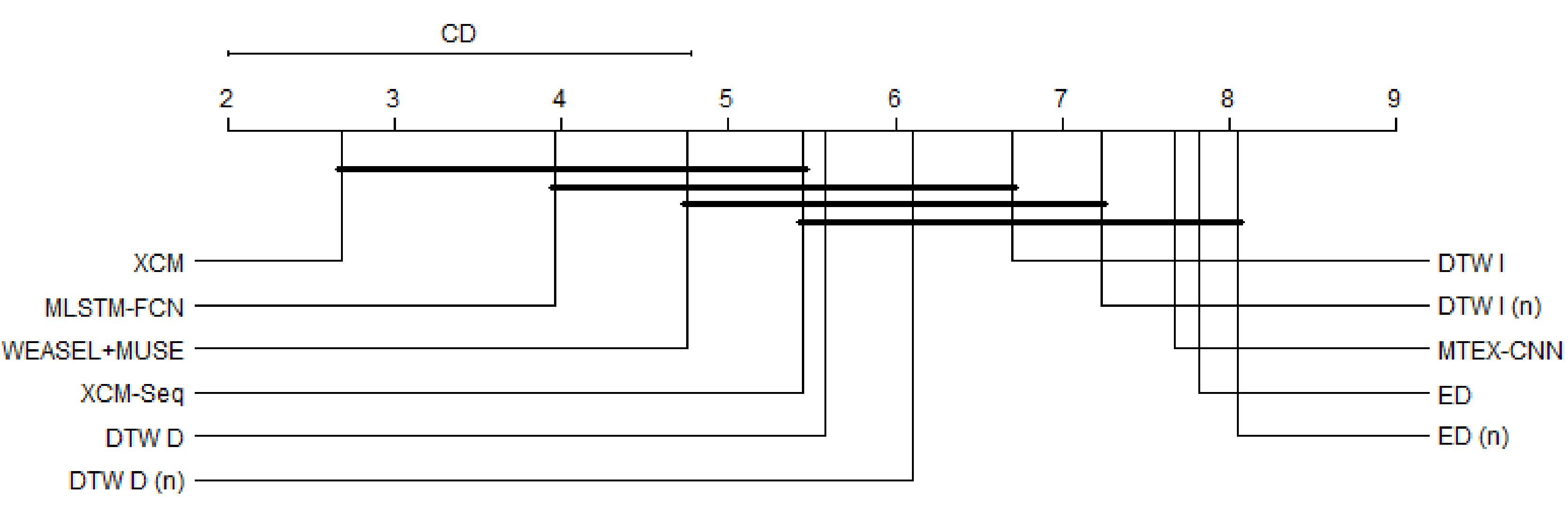}
	\caption{Critical difference plot of the MTS classifiers on the UEA datasets with alpha equal to~0.05.}
	\label{fig:CDPlot_MTS}
\end{figure}

\subsection{Explainability}
\label{sec:res_explainability}
In this section, we illustrate how our approach XCM reconciles performance and explainability and~show that XCM enables a more precise identification of the regions of the input data that are important for predictions compared to the current deep learning MTS classifier also providing faithful explainability---MTEX-CNN. 
We perform the comparison on a synthetic dataset as the construction of such a dataset allows us to know the expected explanations, with such information not being available in the public UEA datasets. Concerning the evaluation of the results, we adopt the intersection-over-union as a metric, i.e.,~the extent of overlap between the predicted and expected~explanations.

The synthetic dataset is composed of 20 MTS (50\%/50\% train/test split) with a length of 100, two dimensions, and two balanced classes. The~difference between the 10 MTS belonging to the negative class and the one belonging to the positive class stems from a 20\% time window of the MTS. Negative class MTS are sine waves, and as~illustrated in the plot on the top part of Figure~\ref{fig:synthetic_MTEX_XCM}, positive class MTS are sine waves with a square signal on 20\% of the dimension 1 (see timestamps between 60 and 80). 

First, MTEX-CNN and XCM ($Batch \, Size$: 1, $Window \, Size$: 20\%) correctly predict the 10 MTS of the test set (accuracy 100\%). We observe that XCM and MTEX-CNN obtain the same performance whereas XCM has around 10 times fewer parameters than MTEX-CNN (trainable parameters: XCM 17k, MTEX-CNN 232k).

\begin{figure}[H]
	\centering
	\includegraphics[width=.99\linewidth]{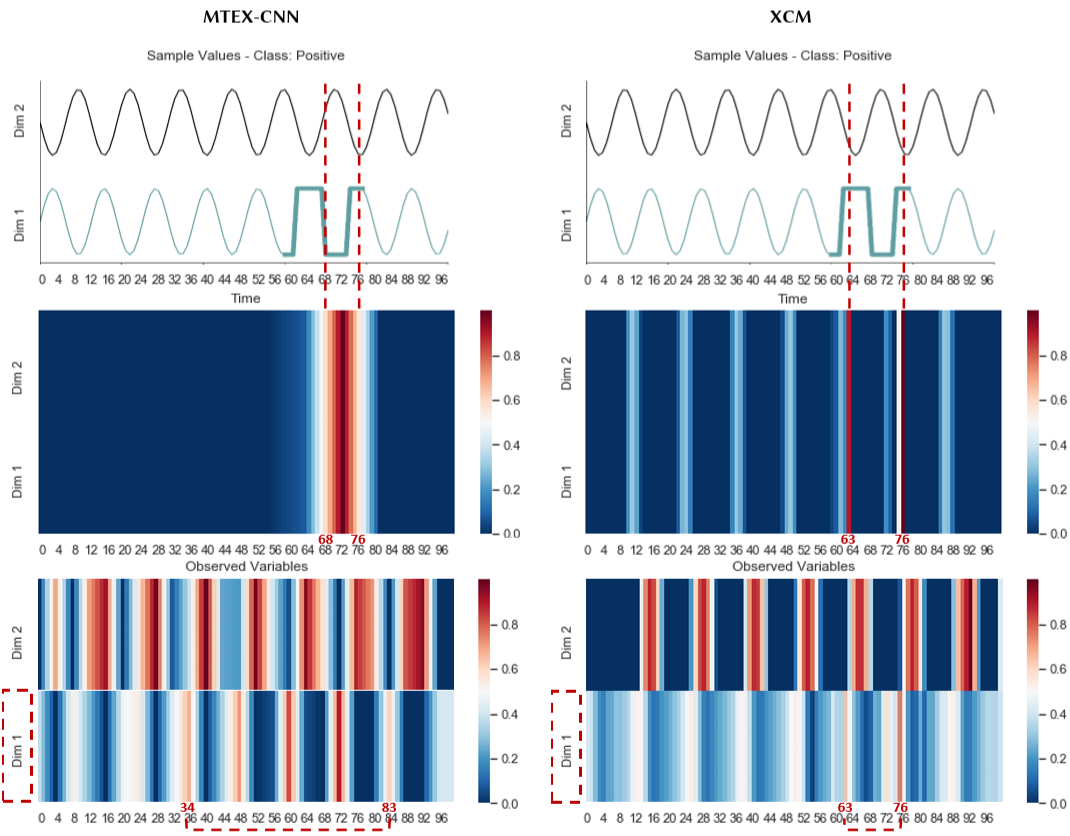}
	\caption{Observed variables and time attribution maps supporting the correct MTEX-CNN and XCM predictions of an MTS from the synthetic dataset belonging to the class \textit{Positive}. Abbreviation: Dim---Dimension.}
	\label{fig:synthetic_MTEX_XCM}
\end{figure}

Moreover, MTEX-CNN and XCM with Grad-CAM all correctly identify the discriminative time window. However, as~shown in Figure~\ref{fig:synthetic_MTEX_XCM}, the~attribution maps of MTEX-CNN and XCM with the same explainability method (Grad-CAM) are different. Figure~\ref{fig:synthetic_MTEX_XCM} shows one MTS sample belonging to the positive class, and~the time and observed variables attribution maps supporting MTEX-CNN and XCM predictions. Attribution maps are heatmaps of the same size as the input data. The~more intense the red, the~stronger the features (observed variables, time) positively contribute to the prediction.
We observe that the attribution maps drawn from XCM are more precise than the ones from MTEX-CNN, i.e.,~the intersection-over-union of the explanations is higher for XCM than for MTEX-CNN (intersection-over-union: XCM 0.65 versus MTEX-CNN 0.4).
On the time attribution map, high attribution values (above 0.6) for XCM begin on timestamp 63 and end on timestamp 76 (expected: $[60,80]$, intersection-over-union: 0.65), whereas for MTEX-CNN they begin later (timestamp 68, intersection-over-union: 0.4). 
Concerning the attribution map of the observed variables, as~expected, we see that high attributions values on the discriminative dimension (dimension 1) appear at the same timestamps as high attribution values on the time attribution map for XCM (timestamps 63 and 76, intersection-over-union: 0.65). Nonetheless, the~observed variables attribution map of MTEX-CNN shows high attribution values on a window larger than the discriminative one (timestamps range $[34,83]$, intersection-over-union: 0.41).
As MTEX-CNN attribution maps exhibit a red color gradient, the~precision of identification of the regions of the input data on MTEX-CNN attribution maps could be enhanced by setting a higher threshold than 0.6 for the attribution values. However, the~red color gradient is due to the upsampling processes needed to match the 2D/1D output features maps of MTEX-CNN to the size of the input data when applying Grad-CAM. Grad-CAM is applied at a local level, which means that we would need to potentially set a different threshold for each instance and that would render MTEX-CNN explainability method impractical. Thus, based on the same attribution threshold (0.6), XCM allows a more precise identification of the regions of the input data that are important for predictions than MTEX-CNN.
Both MTEX-CNN and XCM have periodically high attribution values on dimension 2 of the observed variables attribution maps. It could be surprising as the sinusoidal signal on this dimension is the same across all MTS; however, the fact that this information is uniformly high or low renders it irrelevant for explanations.
Therefore, considering that XCM-Seq attributions maps are the same as XCM ones, we can assume that the use of half padding on the different convolution layers to reduce the number of parameters in MTEX-CNN, i.e., 
 the use of upsampling to retrieve the input data dimensions on the attribution maps, can lead to a less precise identification of the regions of the input data that are important for~predictions.

\subsection{Real-World~Application}
\label{sec:application}
Machine learning methods have great potential to improve the detection of determining events for milk production in dairy farms, which is one of the most important steps toward meeting both food production and environmental goals~\citep{Searchinger18}.
A key factor for dairy farms performance is reproduction.
Reproduction directly impacts milk production as cows start to produce milk after giving birth to a calf; and milk productivity declines after the first 3 months. Furthermore, the~most prevalent reason for cow culling, the~act of slaughtering a cow, is a reproduction issue (e.g., long interval between two calves)~\citep{Bascom98}. Thus, it is crucial to detect estrus, the~only period when the cow is susceptible to pregnancy, to~timely inseminate cows and therefore optimize resource use in dairy~farms. 

The ground truth is estrus estimation using automated progesterone analysis in milk~\citep{Cutullic11,Tenghe15}. However, the~cost of this solution prohibits its extensive implementation. Thus, the~machine learning challenge lies in developing a binary MTS classifier to detect estrus (class estrus/non-estrus) based on affordable sensor data (activity and body temperature). Commercial solutions based on these affordable sensor data have been developed. Nonetheless, their adoption rate remains moderate~\citep{Steenveld15}. These commercial detection solutions suffer from insufficient performance (false alerts and incomplete estrus coverage) and from a lack of justifications supporting alerts. Therefore, aside from an enhanced performance, decision support solutions need to provide to the farmers some explanations supporting the~alerts.

The offline dataset consists of 15.5k MTS samples of length 4 with seven variables: the body temperature variable and six activity variables (\textit{rumination}, \textit{ingestion}, \textit{rest}, \textit{standing up}, \textit{over activity} and \textit{other activity}). A~time series corresponds to a 4 day period (MTS length 4): the day of estrus (\textit{Day 0}) and the previous 3 days. The~labels are set with the ground truth in estrus detection---progesterone dosage in whole milk.
We compare XCM with Grad-CAM to a reference commercial solution (HeatPhone~\citep{Chanvallon14}) and the most accurate state-of-the-art MTS classifier of our benchmark (see Section~\ref{sec:eval_algo}) on this real-world application---MLSTM-FCN---with SHAP~\citep{Lundberg17}. As~far as we have seen, an~architecture concatenating a LSTM network with a CNN such as MLSTM-FCN can only rely on post hoc model-agnostic explainability methods to support its predictions. We chose the state-of-the-art explainability method SHAP as its granularity of explanation is comparable to Grad-CAM (both global and local). Indeed, Grad-CAM can also offer global explainability by averaging the attribution maps values per class. SHAP provides the relative importance of the observed variables and timestamps on predictions. 
Performance is calculated following a five-fold cross-validation and an arithmetic mean of the F1-scores on test sets. The~choice of this metric is driven by two reasons. First, no assumption is made about the dairy management style; farmers can favor a higher estrus detection rate (higher recall) or fewer false alerts (higher precision) according to their needs. Second, there is a class imbalance (33\% of estrus days) which renders irrelevant the accuracy~metric.

As presented in Table~\ref{tab:rw_application}, we observe that XCM outperforms the current state-of-the-art deep learning approach (MLSTM-FCN) and the reference commercial solution by increasing the average F1-score (69.7\% versus 63.1 \% and 55.3\%) and obtaining the lowest variability across folds (1.5\% versus 1.5\% and 5.1\%).
In addition, concerning XCM explainability, Figure~\ref{fig:estrus} shows an example of the observed variables and time attribution maps supporting the correct prediction of an MTS sample belonging to the class \textit{Estrus}. We plot the MTS sample with a heatmap to ease the readability.
The intersection of attribution maps and sample values inform us that the prediction was made mainly based on the presence of a high overactivity (or low rest) of the animal on the day of estrus (attribution values above 0.6 on \textit{Day 0} and on the variable \textit{over activity}, which has a high value). This behavior is aligned with the literature on estrus detection~\citep{Gaillard16}, as it is the behavior associated with most of the~estrus.

\begin{specialtable}[H]
	\widetable
	\caption{Estrus detection F1-score on test sets with 95\% confidence~interval.}
	\centering
	\label{tab:rw_application}
	\setlength{\cellWidtha}{\columnwidth/4-2\tabcolsep-0.5in}
	\setlength{\cellWidthb}{\columnwidth/4-2\tabcolsep-0.1in}
	\setlength{\cellWidthc}{\columnwidth/4-2\tabcolsep-0.1in}
	\setlength{\cellWidthd}{\columnwidth/4-2\tabcolsep+.5in}
	\scalebox{1}[1]{\begin{tabularx}{.95\columnwidth}{>{\PreserveBackslash\raggedright}m{\cellWidtha}>{\PreserveBackslash\centering}m{\cellWidthb}>{\PreserveBackslash\centering}m{\cellWidthc}>{\PreserveBackslash\centering}m{\cellWidthd}}
			\toprule
			\textbf{}&\textbf{XCM}&\textbf{MLSTM-FCN}&\textbf{Commercial Solution}\\
			\midrule
			F1-Score & 69.7 $\pm$ 1.5 & 63.1 $\pm$ 1.5 & 55.3 $\pm$ 5.1 \\
			\bottomrule	
	\end{tabularx}}
\end{specialtable}

The current state-of-the-art MTS classifiers MLSTM-FCN and XCM have different explainability methods (SHAP---post hoc model-agnostic, Grad-CAM---post hoc model-specific) which come with their own form of explanations.
In order to assess and benchmark these two MTS classifiers also with respect to their explainability, we use a framework that we have proposed in~\citep{Fauvel20Framework}. The~framework details a set of characteristics (performance, model comprehensibility, granularity of the explanations, information type, faithfulness and user category) that systematize the performance-explainability assessment of machine learning methods. The~results of the framework are represented in a parallel coordinates plot in Figure~\ref{fig:Framework}.
Both deep learning approaches are hard-to-understand models (Comprehensibility: \textit{Black-Box}) which provide explanations at both global and local levels (Granularity: \textit{Global and Local}) that can be analyzed by a domain expert (User: \textit{Domain Expert}).
However, in~addition to giving the relative importance of observed variables and time as MLSTM-FCN with SHAP, XCM with Grad-CAM provides more informative explanations by supplying the corresponding sample values (Information: MLSTM-FCN with SHAP---\textit{Features+Time} and XCM with Grad-CAM---\textit{Features+Time+Values}).
Furthermore, unlike MLSTM-FCN with SHAP and as discussed in Section~\ref{sec:soa_explainability}, XCM with Grad-CAM approach provides faithful explanations, which is a prerequisite to reduce solution mistrust from the farmers (Faithfulness: MLSTM-FCN with SHAP---\textit{Imperfect} and XCM with Grad-CAM---\textit{Perfect}).
Therefore, XCM outperforms the current state-of-the-art algorithm on the real-world application (Performance: \textit{Best}), while enhancing explainability by providing faithful and more informative~explanations.

\begin{figure}[H]
	\centering
	\includegraphics[width=.7\linewidth]{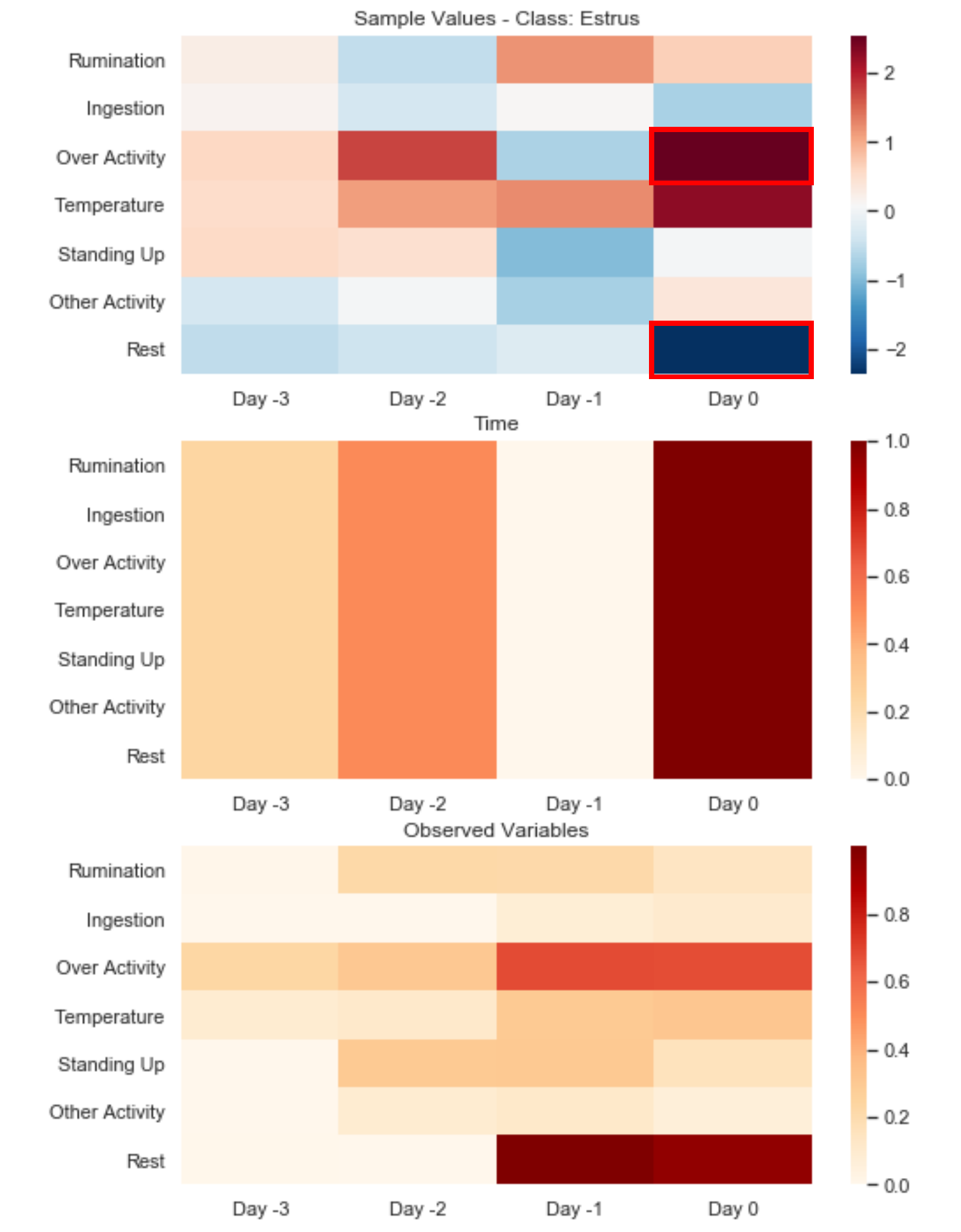}
	\caption{Observed variables and time attribution maps supporting the correct XCM prediction of an MTS from the real-world test set, which belongs to the class \textit{Estrus}. The~MTS sample is represented under the form of a heatmap with the regions important for the prediction highlighted with a red~square.}
	\label{fig:estrus}
\end{figure}

Finally, the~performance-explainability framework introduced in the previous paragraph can also be used to identify the limitations of XCM, which point to the  directions to improve our approach.
We see in Figure~\ref{fig:Framework} that the level of information of the explanation provided by XCM with Grad-CAM (\textit{Features+Time+Values}) could be enhanced.
Therefore, aside from automating the hyperparameter setting of XCM ($Window \, Size$), it would be interesting to work on synthesizing the attribution maps to improve the level of information.

\begin{figure}[H]
	\centering
	\includegraphics[width=\linewidth]{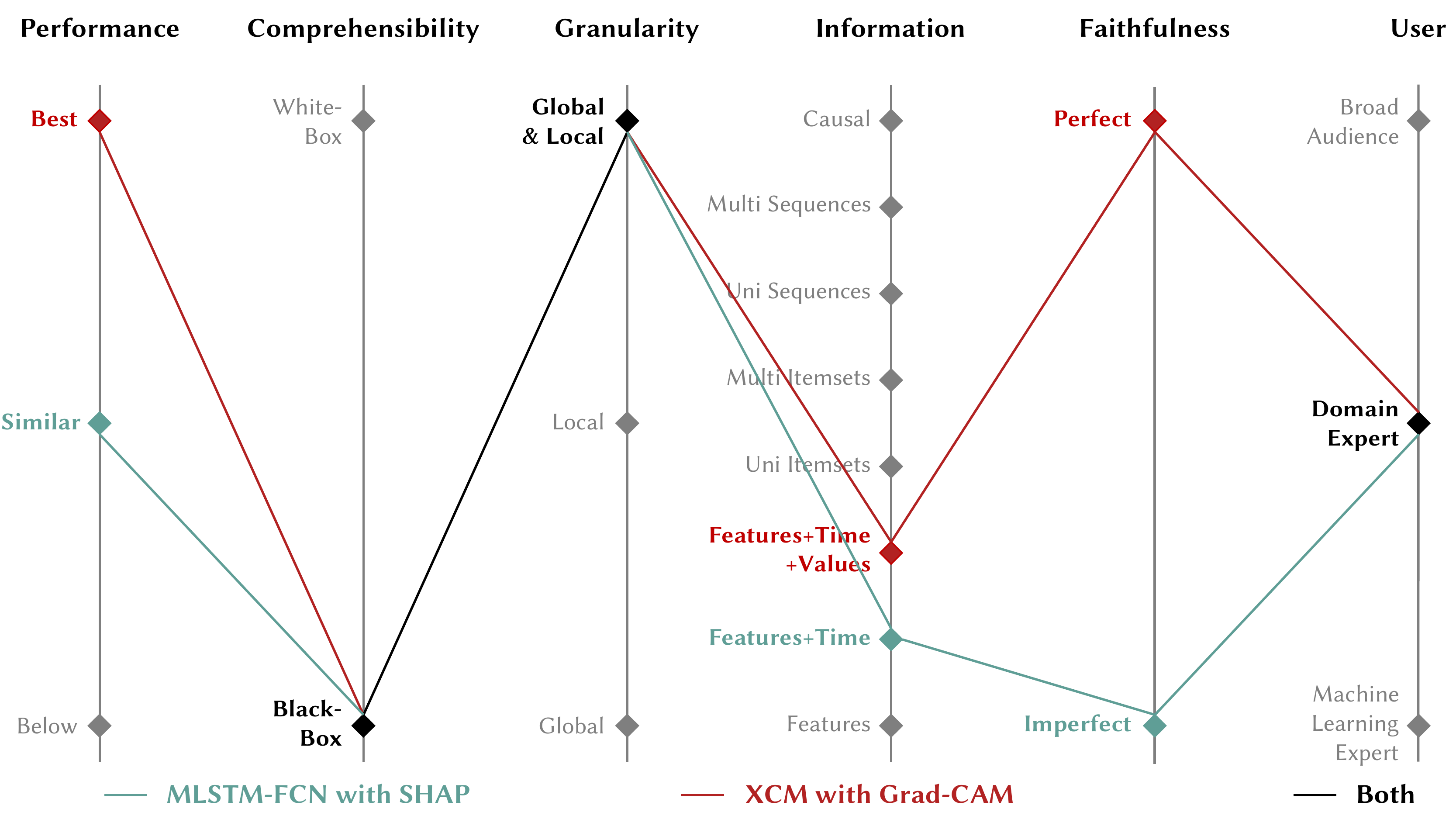}
	\caption{Parallel coordinates plot of XCM and the state-of-the-art MTS classifiers on the real-world application. Performance evaluation method: 5-fold cross-validation and an arithmetic mean of the F1-scores. As~presented in Section~\ref{rw_mts_classifiers}, the~models evaluated in the benchmark are: DTW$_{D}$, DTW$_{I}$, FCN, gRSF, LPS, MLSTM-FCN, MTEX-CNN, mv-ARF, ResNet, SMTS, UFS, WEASEL+MUSE and~XCM.}
	\label{fig:Framework}
\end{figure}

\section{Conclusions}
We have presented XCM, a~new compact and explainable convolutional neural network for MTS classification, which extracts information relative to the observed variables and time directly from the input data.
XCM exhibits a better average rank than the state-of-the-art classifiers on both the large and small public UEA datasets.
Moreover, it was designed to enable faithful explainability based on Grad-CAM method and the precise identification of the regions of the input data that are important for predictions.
Following the illustration of the performance and explainability of XCM on a synthetic dataset, we showed how XCM can outperform the current most accurate state-of-the-art algorithm MLSTM-FCN on a real-world application while enhancing explainability by providing faithful and more informative~explanations.

{\textls[-25]{In our future work, we would like to automate XCM hyperparameter setting ($Window \, Size$)} and evaluate the impact of different fusion methods of the 2D and 1D feature maps (e.g., weighting scheme) on XCM performance.}
With regard to explainability, it would be interesting to further enhance the explanations of XCM with Grad-CAM by synthesizing the attribution maps with multidimensional sequential patterns to improve the level of~information.

\vspace{6pt} 

\authorcontributions{{Conceptualization,  K.F.; methodology, K.F., T.L., V.M., E.F. and A.T.; validation, K.F., V.M., E.F. and A.T.; formal analysis, K.F. and T.L.; investigation, K.F.; data curation, K.F.; writing---original draft preparation, K.F.; writing---review and editing, K.F., V.M., E.F. and A.T.; visualization, K.F.; supervision, V.M., E.F. and A.T.; project administration, E.F. and A.T. All authors have read and agreed to the published version of the manuscript.}
}

\funding{This work was supported by the French National Research Agency under the Investments for the Future Program (ANR-16-CONV-0004), the project Deffilait (ANR-15-CE20-0014) and the Inria Project Lab Hybrid Approaches for Interpretable AI (HyAIAI).}

\institutionalreview{This work was carried out in accordance with the guidelines for animal research of the French Ministry of Agriculture (decret NOR AGRG 1231951D) and approved by the ``Comite National de R\'eflexion Ethique sur l'Experimentation Animale'' (Authorization of the French Ministry of Higher Education, Research and Innovation reference APAFIS 3122-2015112718172611).
}

\informedconsent{{Not applicable} 
}

\dataavailability{The UEA multivariate time series classification archive is available online: \url{https://www.timeseriesclassification.com/index.php} 
} 

\acknowledgments{We would like to thank Philippe Faverdin for his invaluable feedback that has been instrumental for our work.}

\conflictsofinterest{The authors declare no conflict of~interest.} 

\end{paracol}

\reftitle{References}


\begin{thebibliography}{999}

\bibitem[Chen \em{et~al.}(2019)Chen, Bian, Xing, and Liu]{Chen19}
Chen, C.; Bian, J.; Xing, C.; Liu, T.
\newblock Investment {B}ehaviors {C}an {T}ell {W}hat {I}nside: {E}xploring
  {S}tock {I}ntrinsic {P}roperties for {S}tock {T}rend {P}rediction.
\newblock  In Proceedings of the 25th ACM SIGKDD International Conference on
  Knowledge Discovery and Data Mining, Anchorage, USA, August 4--8, 2019.

\bibitem[Li \em{et~al.}(2018)Li, Rong, Meng, Lu, Kwok, and Cheng]{Li18}
Li, J.; Rong, Y.; Meng, H.; Lu, Z.; Kwok, T.; Cheng, H.
\newblock T{ATC}: {P}redicting {A}lzheimer’s {D}isease with {A}ctigraphy
  {D}ata.
\newblock  In Proceedings of the 24th ACM SIGKDD International Conference on
  Knowledge Discovery and Data Mining, London, United Kingdom, August 19--23, 2018.

\bibitem[Jiang \em{et~al.}(2019)Jiang, Song, Huang, Song, Xia, Cai, Wang, Kim,
  and Shibasaki]{Jiang19}
Jiang, R.; Song, X.; Huang, D.; Song, X.; Xia, T.; Cai, Z.; Wang, Z.; Kim, K.;
  Shibasaki, R.
\newblock Deep{U}rban{E}vent: {A} {S}ystem for {P}redicting {C}itywide {C}rowd
  {D}ynamics at {B}ig {E}vents.
\newblock  In Proceedings of the 25th ACM SIGKDD International Conference on
  Knowledge Discovery and Data Mining, Anchorage, USA, August 4--8, 2019.

\bibitem[Fauvel \em{et~al.}(2020)Fauvel, Balouek-Thomert, Melgar, Silva,
  Simonet, Antoniu, Costan, Masson, Parashar, Rodero, and Termier]{Fauvel20}
Fauvel, K.; Balouek-Thomert, D.; Melgar, D.; Silva, P.; Simonet, A.; Antoniu,
  G.; Costan, A.; Masson, V.; Parashar, M.; Rodero, I.; Termier, A.
\newblock A {D}istributed {M}ulti-{S}ensor {M}achine {L}earning {A}pproach to
  {E}arthquake {E}arly {W}arning.
\newblock  In Proceedings of the 34th AAAI Conference on Artificial
  Intelligence, New York, USA, February 7--12, 2020.

\bibitem[Karim \em{et~al.}(2019)Karim, Majumdar, Darabi, and Harford]{Karim19}
Karim, F.; Majumdar, S.; Darabi, H.; Harford, S.
\newblock Multivariate {LSTM-FCN}s for {T}ime {S}eries {C}lassification.
\newblock {\em Neural Netw.} {\bf 2019}, {\em 116},~237--245.

\bibitem[Sch{\"a}fer and Leser(2017)]{Schafer17}
Sch{\"a}fer, P.; Leser, U.
\newblock Multivariate {T}ime {S}eries {C}lassification with {WEASEL + MUSE}.
\newblock {arXiv, 2017}.

\bibitem[Bagnall \em{et~al.}(2018)Bagnall, Lines, and Keogh]{Bagnall18}
Bagnall, A.; Lines, J.; Keogh, E.
\newblock {The {UEA} {M}ultivariate {T}ime {S}eries {C}lassification {A}rchive {2018}.}
\newblock arXiv, 2018. 

\bibitem[Sch\"{a}fer and H\"{o}gqvist(2012)]{Schafer12}
Sch\"{a}fer, P.; H\"{o}gqvist, M.
\newblock {SFA}: {A} {S}ymbolic {F}ourier {A}pproximation and {I}ndex for
  {S}imilarity {S}earch in {H}igh {D}imensional {D}atasets.
\newblock  In Proceedings of the 15th International Conference on Extending
  Database Technology, Berlin, Germany, March 27--30, 2012; pp. 516--527.

\bibitem[Rudin(2019)]{Rudin19}
Rudin, C.
\newblock Stop {E}xplaining {B}lack {B}ox {M}achine {L}earning {M}odels for
  {H}igh {S}takes {D}ecisions and {U}se {I}nterpretable {M}odels {I}nstead.
\newblock {\em Nat. Mach. Intell.} {\bf 2019}, {\em 1},~206--215.

\bibitem[Selvaraju \em{et~al.}(2019)Selvaraju, Das, Vedantam, Cogswell, Parikh,
  and Batra]{Selvaraju19}
Selvaraju, R.; Das, A.; Vedantam, R.; Cogswell, M.; Parikh, D.; Batra, D.
\newblock Grad-{CAM}: {V}isual {E}xplanations from {D}eep {N}etworks via
  {G}radient-{B}ased {L}ocalization.
\newblock {\em Int. J. Comput. Vis.} {\bf 2019}, {\em
  128},~336--359.

\bibitem[Adebayo \em{et~al.}(2018)Adebayo, Gilmer, Muelly, Goodfellow, Hardt,
  and Kim]{Adebayo18}
Adebayo, J.; Gilmer, J.; Muelly, M.; Goodfellow, I.; Hardt, M.; Kim, B.
\newblock Sanity {C}hecks for {S}aliency {M}aps.
\newblock  In Proceedings of the 32nd International Conference on Neural
  Information Processing Systems, Montreal, Canada, December 3--8, 2018.

\bibitem[Assaf \em{et~al.}(2019)Assaf, Giurgiu, Bagehorn, and
  Schumann]{Assaf19}
Assaf, R.; Giurgiu, I.; Bagehorn, F.; Schumann, A.
\newblock M{TEX-CNN}: {M}ultivariate {T}ime {S}eries {EX}planations for
  {P}redictions with {C}onvolutional {N}eural {N}etworks.
\newblock  In Proceedings of the IEEE International Conference on Data Mining,
  Beijing, China, November 8--11, 2019.

\bibitem[Goodfellow \em{et~al.}(2016)Goodfellow, Bengio, and
  Courville]{Goodfellow16}
Goodfellow, I.; Bengio, Y.; Courville, A.
\newblock {\em Deep {L}earning}; MIT Press: Cambridge, MA, USA,
 2016.

\bibitem[Huang \em{et~al.}(2017)Huang, Liu, {Van Der Maaten}, and
  Weinberger]{Huang17}
Huang, G.; Liu, Z.; {Van Der Maaten}, L.; Weinberger, K.
\newblock Densely {C}onnected {C}onvolutional {N}etworks.
\newblock  In Proceedings of the 2017 IEEE Conference on Computer Vision and Pattern Recognition,
  Hawaii, USA, July 21--26, 2017.

\bibitem[Sutskever \em{et~al.}(2014)Sutskever, Vinyals, and Le]{Sutskever14}
Sutskever, I.; Vinyals, O.; Le, Q.
\newblock Sequence to {S}equence {L}earning with {N}eural {N}etworks.
\newblock  In Proceedings of the 27th International Conference on Neural
  Information Processing Systems, Montreal, Canada, December 8--13, 2014.

\bibitem[Devlin \em{et~al.}(2019)Devlin, Chang, Lee, and Toutanova]{Devlin19}
Devlin, J.; Chang, M.; Lee, K.; Toutanova, K.
\newblock B{ERT}: {P}re-{T}raining of {D}eep {B}idirectional {T}ransformers for
  {L}anguage {U}nderstanding.
\newblock  In Proceedings of the 2019 Conference of the North {A}merican Chapter of the Association for Computational Linguistics, 
Minneapolis, USA, June 2--7, 2019.

\bibitem[{Cristian Borges Gamboa}(2017)]{Cristian17}
{Cristian Borges Gamboa}, J.
\newblock Deep {L}earning for {T}ime-{S}eries {A}nalysis.
\newblock arXiv, 2017.

\bibitem[Seto \em{et~al.}(2015)Seto, Zhang, and Zhou]{Seto15}
Seto, S.; Zhang, W.; Zhou, Y.
\newblock Multivariate {T}ime {S}eries {C}lassification {U}sing {D}ynamic
  {T}ime {W}arping {T}emplate {S}election for {H}uman {A}ctivity {R}ecognition.
\newblock In Proceedings of the IEEE Symposium Series on Computational
  Intelligence, Cape Town, South Africa, December 7--10, 2015.
  
\bibitem[Vidal \em{et~al.}(1985)Vidal, Casacuberta, and Segovia]{Vidal85}
Vidal, E.; Casacuberta, F.; Segovia, H.
\newblock Is the {DTW} ``{D}istance'' {R}eally a {M}etric? {A}n {A}lgorithm
  {R}educing the {N}umber of {DTW} {C}omparisons in {I}solated {W}ord
  {R}ecognition.
\newblock {\em Speech Commun.} {\bf 1985}, {\em 4},~333--344.

\bibitem[Shokoohi-Yekta \em{et~al.}(2017)Shokoohi-Yekta, Hu, Jin, Wang, and
  Keogh]{Shokoohi17}
Shokoohi-Yekta, M.; Hu, B.; Jin, H.; Wang, J.; Keogh, E.
\newblock Generalizing {DTW} to the {M}ulti-{D}imensional {C}ase {R}equires an
  {A}daptive {A}pproach.
\newblock {\em Data Min. Knowl. Discov.} {\bf 2017}, {\em
  31},~1–31.

\bibitem[Karlsson \em{et~al.}(2016)Karlsson, Papapetrou, and
  Bostr{\"o}m]{Karlsson16}
Karlsson, I.; Papapetrou, P.; Bostr{\"o}m, H.
\newblock Generalized {R}andom {S}hapelet {F}orests.
\newblock {\em Data Min. Knowl. Discov.} {\bf 2016}, {\em
  30},~1053--1085.

\bibitem[Wistuba \em{et~al.}(2015)Wistuba, Grabocka, and
  Schmidt{-}Thieme]{Wistuba15}
Wistuba, M.; Grabocka, J.; Schmidt{-}Thieme, L.
\newblock Ultra-{F}ast {S}hapelets for {T}ime {S}eries {C}lassification.
\newblock arXiv, 2015.

\bibitem[Baydogan and Runger(2016)]{Baydogan16}
Baydogan, M.; Runger, G.
\newblock Time {S}eries {R}epresentation and {S}imilarity {B}ased on {L}ocal
  {A}utopatterns.
\newblock {\em Data Min. Knowl. Discov.} {\bf 2016}, {\em
  30},~476--509.

\bibitem[Tuncel and Baydogan(2018)]{Tuncel18}
Tuncel, K.; Baydogan, M.
\newblock Autoregressive {F}orests for {M}ultivariate {T}ime {S}eries
  {M}odeling.
\newblock {\em Pattern Recognit.} {\bf 2018}, {\em 73},~202--215.

\bibitem[Baydogan and Runger(2014)]{Baydogan14}
Baydogan, M.; Runger, G.
\newblock Learning a {S}ymbolic {R}epresentation for {M}ultivariate {T}ime
  {S}eries {C}lassification.
\newblock {\em Data Min. Knowl. Discov.} {\bf 2014}, {\em
  29},~400--422.

\bibitem[Wang \em{et~al.}(2017)Wang, Yan, and Oates]{Wang17}
Wang, Z.; Yan, W.; Oates, T.
\newblock Time {S}eries {C}lassification from {S}cratch with {D}eep {N}eural
  {N}etworks: {A} {S}trong {B}aseline.
\newblock In Proceedings of the 2017 International Joint Conference on Neural
  Networks, Anchorage, USA, May 14--19, 2017.

\bibitem[He \em{et~al.}(2016)He, Zhang, Ren, and Sun]{He16}
He, K.; Zhang, X.; Ren, S.; Sun, J.
\newblock Deep {R}esidual {L}earning for {I}mage {R}ecognition.
\newblock In Proceedings of the 2016 IEEE Conference on Computer Vision and
  Pattern Recognition, Las Vegas, USA, June 27--30, 2016.

\bibitem[Zhang \em{et~al.}(2020)Zhang, Gao, Lin, and Lu]{Zhang20}
Zhang, X.; Gao, Y.; Lin, J.; Lu, C.
\newblock TapNet: Multivariate Time Series Classification with Attentional
  Prototypical Network.
\newblock In Proceedings of the 34th {AAAI} Conference on Artificial
  Intelligence, New York, USA, February 7--12, 2020.

\bibitem[Zerveas \em{et~al.}(2021)Zerveas, Jayaraman, Patel, Bhamidipaty, and
  Eickhoff]{Zerveas21}
Zerveas, G.; Jayaraman, S.; Patel, D.; Bhamidipaty, A.; Eickhoff, C.
\newblock A {T}ransformer-{B}ased {F}ramework for {M}ultivariate {T}ime
  {S}eries {R}epresentation {L}earning.
\newblock In Proceedings of the 27th ACM SIGKDD Conference on Knowledge Discovery
  and Data Mining, Virtual Event, August 14--18, 2021.

\bibitem[Du \em{et~al.}(2020)Du, Liu, and Hu]{Du20}
Du, M.; Liu, N.; Hu, X.
\newblock Techniques for {I}nterpretable {M}achine {L}earning.
\newblock {\em Commun. ACM} {\bf 2020}, \emph{63}, 68--77. 

\bibitem[Ribeiro \em{et~al.}(2016)Ribeiro, Singh, and Guestrin]{Ribeiro16}
Ribeiro, M.; Singh, S.; Guestrin, C.
\newblock “{W}hy {S}hould {I} {T}rust {Y}ou?”: {E}xplaining the
  {P}redictions of {A}ny {C}lassifier.
\newblock In Proceedings of the 22nd ACM SIGKDD International Conference on
  Knowledge Discovery and Data Mining, San Francisco, USA, August 13--17, 2016.

\bibitem[Lundberg and Lee(2017)]{Lundberg17}
Lundberg, S.; Lee, S.
\newblock A {U}nified {A}pproach to {I}nterpreting {M}odel {P}redictions.
\newblock In Proceedings of the 31st International Conference on Neural
  Information Processing Systems, Long Beach, USA, December 4--9, 2017.

\bibitem[Ribeiro \em{et~al.}(2018)Ribeiro, Singh, and Guestrin]{Ribeiro18}
Ribeiro, M.; Singh, S.; Guestrin, C.
\newblock Anchors: {H}igh-{P}recision {M}odel-{A}gnostic {E}xplanations.
\newblock In Proceedings of the 32nd AAAI Conference on Artificial
  Intelligence, New Orleans, USA, February 2--7, 2018.

\bibitem[Guidotti \em{et~al.}(2019)Guidotti, Monreale, Giannotti, Pedreschi,
  Ruggieri, and Turini]{Guidotti19}
Guidotti, R.; Monreale, A.; Giannotti, F.; Pedreschi, D.; Ruggieri, S.; Turini,
  F.
\newblock Factual and {C}ounterfactual {E}xplanations for {B}lack {B}ox
  {D}ecision {M}aking.
\newblock {\em IEEE Intell. Syst.} {\bf 2019}, {\em 34},~14--23.

\bibitem[Ancona \em{et~al.}(2018)Ancona, Ceolini, {\"O}ztireli, and
  Gross]{Ancona18}
Ancona, M.; Ceolini, E.; {\"O}ztireli, C.; Gross, M.
\newblock Towards {B}etter {U}nderstanding of {G}radient-{B}ased {A}ttribution
  {M}ethods for {D}eep {N}eural {N}etworks.
\newblock In Proceedings of the International Conference on Learning
  Representations, Vancouver, Canada, May 1--3, 2018.

\bibitem[Erhan \em{et~al.}(2009)Erhan, Bengio, Courville, and Vincent]{Erhan09}
Erhan, D.; Bengio, Y.; Courville, A.; Vincent, P.
\newblock Visualizing {H}igher-{L}ayer {F}eatures of a {D}eep {N}etwork.
\newblock  In Proceedings of the ICML Workshop on Learning Feature Hierarchies,
  Montreal, Canada, June 9, 2009.

\bibitem[Springenberg \em{et~al.}(2015)Springenberg, Dosovitskiy, Brox, and
  Riedmiller]{Springenberg15}
Springenberg, J.; Dosovitskiy, A.; Brox, T.; Riedmiller, M.
\newblock Striving for {S}implicity: {T}he {A}ll {C}onvolutional {N}et.
\newblock In Proceedings of the International Conference on Learning
  Representations (Workshop Track), San Diego, USA, May 7--9, 2015.

\bibitem[Bach \em{et~al.}(2015)Bach, Binder, Montavon, Klauschen, M{\"u}ller,
  and Samek]{Bach15}
Bach, S.; Binder, A.; Montavon, G.; Klauschen, F.; M{\"u}ller, K.; Samek, W.
\newblock On {P}ixel-{W}ise {E}xplanations for {N}on-{L}inear {C}lassifier
  {D}ecisions by {L}ayer-{W}ise {R}elevance {P}ropagation.
\newblock {\em PLoS ONE} {\bf 2015}, {\em 10},~e0130140.

\bibitem[Shrikumar \em{et~al.}(2016)Shrikumar, Greenside, Shcherbina, and
  Kundaje]{Shrikumar16}
{Shrikumar, A.; Greenside, P.; Shcherbina, A.; Kundaje, A.}
\newblock Not {J}ust a {B}lack {B}ox: {L}earning {I}mportant {F}eatures
  {T}hrough {P}ropagating {A}ctivation {D}ifferences.
 \newblock arXiv, 2016.

\bibitem[Sundararajan \em{et~al.}(2017)Sundararajan, Taly, and
  Yan]{Sundararajan17}
Sundararajan, M.; Taly, A.; Yan, Q.
\newblock Axiomatic {A}ttribution for {D}eep {N}etworks.
\newblock  In Proceedings of the 34th International Conference on Machine
  Learning, Sydney, Australia, August 6--11, 2017.

\bibitem[Shrikumar \em{et~al.}(2017)Shrikumar, Greenside, and
  Kundaje]{Shrikumar17}
{Shrikumar, A.; Greenside, P.; Kundaje, A.}
\newblock Learning {I}mportant {F}eatures through {P}ropagating {A}ctivation
  {D}ifferences.
\newblock In Proceedings of the 34th International Conference on Machine
  Learning, Sydney, Australia, August 6--11, 2017.

\bibitem[Lin \em{et~al.}(2014)Lin, Chen, and Yan]{Lin14}
Lin, M.; Chen, Q.; Yan, S.
\newblock Network in {N}etwork.
\newblock arXiv, 2014.

\bibitem[Ioffe and Szegedy(2015)]{Ioffe15}
Ioffe, S.; Szegedy, C.
\newblock Batch {N}ormalization: {A}ccelerating {D}eep {N}etwork {T}raining by
  {R}educing {I}nternal {C}ovariate {S}hift.
\newblock In Proceedings of the 32nd International
  Conference on Machine Learning, Lille, France, July 6--11, 2015.

\bibitem[Nair and Hinton(2010)]{Nair10}
Nair, V.; Hinton, G.
\newblock Rectified {L}inear {U}nits {I}mprove {R}estricted {B}oltzmann
  {M}achines.
\newblock In Proceedings of the 27th International
  Conference on Machine Learning, Haifa, Israel, June 21--24, 2010.

\bibitem[Bjorck \em{et~al.}(2018)Bjorck, Gomes, Selman, and
  Weinberger]{Bjorck18}
Bjorck, N.; Gomes, C.; Selman, B.; Weinberger, K.
\newblock Understanding {B}atch {N}ormalization.
\newblock In Proceedings of the 32nd International Conference on Neural
  Information Processing Systems, Montreal, Canada, December 3--8, 2018.

\bibitem[Szegedy \em{et~al.}(2015)Szegedy, Liu, Jia, Sermanet, Reed, Anguelov,
  Erhan, Vanhoucke, and Rabinovich]{Szegedy15}
Szegedy, C.; Liu, W.; Jia, Y.; Sermanet, P.; Reed, S.; Anguelov, D.; Erhan, D.;
  Vanhoucke, V.; Rabinovich, A.
\newblock Going {D}eeper with {C}onvolutions.
\newblock In Proceedings of the IEEE Conference on Computer Vision and Pattern
  Recognition, Boston, USA, June 7--12, 2015.

\bibitem[Dem\v{s}ar(2006)]{Demsar06}
Dem\v{s}ar, J.
\newblock Statistical {C}omparisons of {C}lassifiers over {M}ultiple {D}ata
  {S}ets.
\newblock {\em J. Mach. Learn. Res.} {\bf 2006}, {\em
  7},~1--30.

\bibitem[Searchinger \em{et~al.}(2018)Searchinger, Waite, Hanson, Ranganathan,
  Dumas, and Matthews]{Searchinger18}
Searchinger, T.; Waite, R.; Hanson, C.; Ranganathan, J.; Dumas, P.; Matthews,
  E.
\newblock {\em Creating a Sustainable Food Future}; World Resources Institute: Washington, USA,
  2018.

\bibitem[Bascom and Young(1998)]{Bascom98}
Bascom, S.; Young, A.
\newblock A {S}ummary of the {R}easons {W}hy {F}armers {C}ull {C}ows.
\newblock {\em J. Dairy Sci.} {\bf 1998}, {\em 81}, 2299--2305.

\bibitem[Cutullic \em{et~al.}(2011)Cutullic, Delaby, Gallard, and
  Disenhaus]{Cutullic11}
Cutullic, E.; Delaby, L.; Gallard, Y.; Disenhaus, C.
\newblock Dairy {C}ows' {R}eproductive {R}esponse to {F}eeding {L}evel
  {D}iffers {A}ccording to the {R}eproductive {S}tage and the {B}reed.
\newblock {\em Animal} {\bf 2011}, {\em 5}, 731--740.

\bibitem[Tenghe \em{et~al.}(2015)Tenghe, Bouwman, Berglund, Strandberg, Blom,
  and Veerkamp]{Tenghe15}
Tenghe, A.; Bouwman, A.; Berglund, B.; Strandberg, E.; Blom, J.; Veerkamp, R.
\newblock Estimating genetic parameters for fertility in dairy cows from
  in-line milk progesterone profiles.
\newblock {\em J. Dairy Sci.} {\bf 2015}, {\em 98},~5763--5773.

\bibitem[Steeneveld and Hogeveen(2015)]{Steenveld15}
Steeneveld, W.; Hogeveen, H.
\newblock Characterization of {D}utch {D}airy {F}arms {U}sing {S}ensor
  {S}ystems for {C}ow {M}anagement.
\newblock {\em J. Dairy Sci.} {\bf 2015}, {\em 98},~709--717.

\bibitem[Chanvallon \em{et~al.}(2014)Chanvallon, Coyral-Castel, Gatien, Lamy,
  Ribaud, Allain, Cl\'ement, and Salvetti]{Chanvallon14}
Chanvallon, A.; Coyral-Castel, S.; Gatien, J.; Lamy, J.; Ribaud, D.; Allain,
  C.; Cl\'ement, P.; Salvetti, P.
\newblock Comparison of {T}hree {D}evices for the {A}utomated {D}etection of
  {E}strus in {D}airy {C}ows.
\newblock {\em Theriogenology} {\bf 2014}, {\em 82},~734--741.

\bibitem[Gaillard \em{et~al.}(2016)Gaillard, Barbu, S{\o}rensen, Sehested,
  Callesen, and Vestergaard]{Gaillard16}
Gaillard, C.; Barbu, H.; S{\o}rensen, M.; Sehested, J.; Callesen, H.;
  Vestergaard, M.
\newblock Milk {Y}ield and {E}strous {B}ehavior {D}uring {E}ight {C}onsecutive
  {E}struses in {H}olstein {C}ows {F}ed {S}tandardized or {H}igh {E}nergy
  {D}iets and {G}rouped {A}ccording to {L}ive {W}eight {C}hanges in {E}arly
  {L}actation.
\newblock {\em J. Dairy Sci.} {\bf 2016}, {\em 99},~3134--3143.

\bibitem[Fauvel \em{et~al.}(2020)Fauvel, Masson, and
  Fromont]{Fauvel20Framework}
Fauvel, K.; Masson, V.; Fromont, {\'E}.
\newblock A {P}erformance-{E}xplainability {F}ramework to {B}enchmark {M}achine
  {L}earning {M}ethods: {A}pplication to {M}ultivariate {T}ime {S}eries
  {C}lassifiers.
\newblock In Proceedings of the IJCAI-PRICAI 2020 Workshop on Explainable AI, Virtual Event, January 8, 2021.
 
\end{thebibliography}
\end{document}